%% file: main.tex
\documentclass[sigconf]{acmart}

\AtBeginDocument{%
  \providecommand\BibTeX{{%
    \normalfont B\kern-0.5em{\scshape i\kern-0.25em b}\kern-0.8em\TeX}}}

\copyrightyear{2022}
\acmYear{2022}
\setcopyright{rightsretained}
\acmConference[SIGGRAPH '22 Conference Proceedings]{Special Interest Group on Computer Graphics and Interactive Techniques Conference Proceedings}{August 7--11, 2022}{Vancouver, BC, Canada}
\acmBooktitle{Special Interest Group on Computer Graphics and Interactive Techniques Conference Proceedings (SIGGRAPH '22 Conference Proceedings), August 7--11, 2022, Vancouver, BC, Canada}
\acmDOI{10.1145/3528233.3530701}
\acmISBN{978-1-4503-9337-9/22/08}

\input{rgb}

\input{defs}

\usepackage{float}
\usepackage{subcaption}

\usepackage{calc}
\usepackage{multirow}
\usepackage{enumitem}

\usepackage[ruled]{algorithm2e} 

\SetAlFnt{\small}
\SetAlCapFnt{\small}
\SetAlCapNameFnt{\small}
\SetAlCapHSkip{0pt}

\newcommand{\revise}[1]{\textcolor{black}{#1}} %
\newcommand{\newrevise}[1]{\textcolor{black}{#1}} 

\usepackage[noend]{algorithmic}

\begin{document}

\title{NeuralPassthrough: Learned Real-Time View Synthesis for VR}

\author{Lei Xiao}
\affiliation{%
  \institution{Reality Labs Research, Meta}
  \city{}
  \country{United States of America}
}
\email{lei.xiao@fb.com}

\author{Salah Nouri}
\affiliation{%
  \institution{Reality Labs Research, Meta}
  \city{}
  \country{United States of America}
}
\email{snouri@fb.com}

\author{Joel Hegland}
\affiliation{%
  \institution{Reality Labs Research, Meta}
  \city{}
  \country{United States of America}
}
\email{hegland@fb.com}

\author{Alberto Garcia Garcia}
\affiliation{%
  \institution{Reality Labs, Meta}
  \city{}
  \country{Switzerland}
}
\email{agarciagarcia@fb.com}

\author{Douglas Lanman}
\affiliation{%
 \institution{Reality Labs Research, Meta}
 \city{}
 \country{United States of America}
}
\email{douglas.lanman@fb.com}

\renewcommand{\shortauthors}{Xiao, L. et al}

\begin{abstract}
Virtual reality (VR) headsets provide an immersive, stereoscopic visual experience, but at the cost of blocking users from directly observing their physical environment. Passthrough techniques are intended to address this limitation by leveraging outward-facing cameras to reconstruct the images that would otherwise be seen by the user without the headset. This is inherently a real-time view synthesis challenge, since passthrough cameras cannot be physically co-located with the user's eyes. Existing passthrough techniques suffer from distracting reconstruction artifacts, largely due to the lack of accurate depth information (especially for near-field and disoccluded objects), and also exhibit limited image quality (e.g., being low resolution and monochromatic). In this paper, we propose the first {\em{learned}} passthrough method and assess its performance using a custom VR headset that contains a stereo pair of RGB cameras. Through both simulations and experiments, we demonstrate that our learned passthrough method delivers superior image quality compared to state-of-the-art methods, while meeting strict VR requirements for real-time, perspective-correct stereoscopic view synthesis over a wide field of view for desktop-connected headsets.
\end{abstract}

\ccsdesc[500]{Computing methodologies~Mixed / augmented reality}
\ccsdesc[500]{Computing methodologies~Machine learning}
\ccsdesc[500]{Computing methodologies~Image-based rendering}

%
%

\keywords{Passthrough, Real-Time View Synthesis, Virtual Reality} 

\begin{teaserfigure}
\centering
\includegraphics[width=\columnwidth]{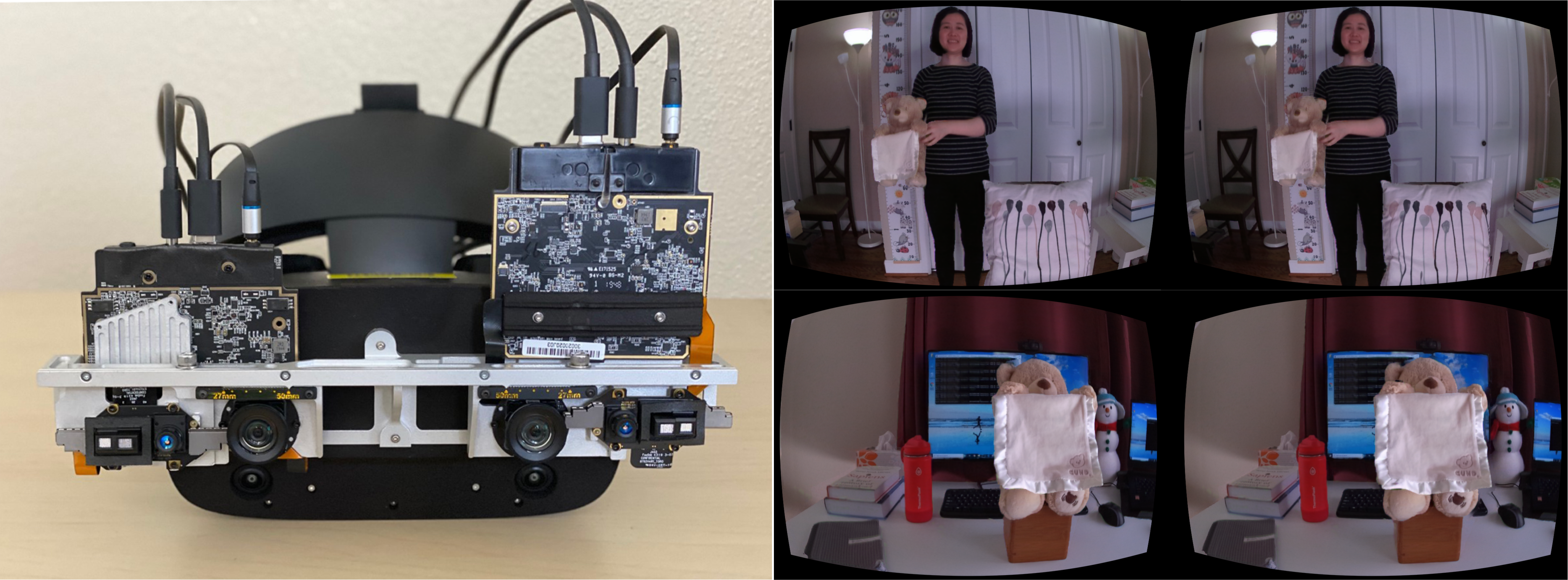}
\caption{We demonstrate the first neural view synthesis method that is optimized to meet the unique requirements for VR passthrough, synthesizing perspective-correct viewpoints in real time and with high visual fidelity. \emph{Left}: We demonstrate performance using a custom-built VR headset, containing a stereo RGB camera rig with an adjustable baseline. \emph{Right}: Our method runs in real-time and supports dynamic scenes ({{top}}) and near-field objects ({{bottom}}).}
\label{fig:teaser}
\end{teaserfigure}

\maketitle

\input{introduction}

\input{previous_work}

\input{method}

\input{results}

\input{limitations}

\input{conclusion}


\bibliographystyle{ACM-Reference-Format}
\bibliography{bibliography}

\end{document}

%% file: rgb.tex
\definecolor{AliceBlue}{rgb}{0.94,0.97,1.00}
\definecolor{AntiqueWhite1}{rgb}{1.00,0.94,0.86}
\definecolor{AntiqueWhite2}{rgb}{0.93,0.87,0.80}
\definecolor{AntiqueWhite3}{rgb}{0.80,0.75,0.69}
\definecolor{AntiqueWhite4}{rgb}{0.55,0.51,0.47}
\definecolor{AntiqueWhite}{rgb}{0.98,0.92,0.84}
\definecolor{BlanchedAlmond}{rgb}{1.00,0.92,0.80}
\definecolor{BlueViolet}{rgb}{0.54,0.17,0.89}
\definecolor{CadetBlue1}{rgb}{0.60,0.96,1.00}
\definecolor{CadetBlue2}{rgb}{0.56,0.90,0.93}
\definecolor{CadetBlue3}{rgb}{0.48,0.77,0.80}
\definecolor{CadetBlue4}{rgb}{0.33,0.53,0.55}
\definecolor{CadetBlue}{rgb}{0.37,0.62,0.63}
\definecolor{CornflowerBlue}{rgb}{0.39,0.58,0.93}
\definecolor{DarkBlue}{rgb}{0.00,0.00,0.55}
\definecolor{DarkCyan}{rgb}{0.00,0.55,0.55}
\definecolor{DarkGoldenrod1}{rgb}{1.00,0.73,0.06}
\definecolor{DarkGoldenrod2}{rgb}{0.93,0.68,0.05}
\definecolor{DarkGoldenrod3}{rgb}{0.80,0.58,0.05}
\definecolor{DarkGoldenrod4}{rgb}{0.55,0.40,0.03}
\definecolor{DarkGoldenrod}{rgb}{0.72,0.53,0.04}
\definecolor{DarkGray}{rgb}{0.66,0.66,0.66}
\definecolor{DarkGreen}{rgb}{0.00,0.39,0.00}
\definecolor{DarkGrey}{rgb}{0.66,0.66,0.66}
\definecolor{DarkKhaki}{rgb}{0.74,0.72,0.42}
\definecolor{DarkMagenta}{rgb}{0.55,0.00,0.55}
\definecolor{DarkOliveGreen1}{rgb}{0.79,1.00,0.44}
\definecolor{DarkOliveGreen2}{rgb}{0.74,0.93,0.41}
\definecolor{DarkOliveGreen3}{rgb}{0.64,0.80,0.35}
\definecolor{DarkOliveGreen4}{rgb}{0.43,0.55,0.24}
\definecolor{DarkOliveGreen}{rgb}{0.33,0.42,0.18}
\definecolor{DarkOrange1}{rgb}{1.00,0.50,0.00}
\definecolor{DarkOrange2}{rgb}{0.93,0.46,0.00}
\definecolor{DarkOrange3}{rgb}{0.80,0.40,0.00}
\definecolor{DarkOrange4}{rgb}{0.55,0.27,0.00}
\definecolor{DarkOrange}{rgb}{1.00,0.55,0.00}
\definecolor{DarkOrchid1}{rgb}{0.75,0.24,1.00}
\definecolor{DarkOrchid2}{rgb}{0.70,0.23,0.93}
\definecolor{DarkOrchid3}{rgb}{0.60,0.20,0.80}
\definecolor{DarkOrchid4}{rgb}{0.41,0.13,0.55}
\definecolor{DarkOrchid}{rgb}{0.60,0.20,0.80}
\definecolor{DarkRed}{rgb}{0.55,0.00,0.00}
\definecolor{DarkSalmon}{rgb}{0.91,0.59,0.48}
\definecolor{DarkSeaGreen1}{rgb}{0.76,1.00,0.76}
\definecolor{DarkSeaGreen2}{rgb}{0.71,0.93,0.71}
\definecolor{DarkSeaGreen3}{rgb}{0.61,0.80,0.61}
\definecolor{DarkSeaGreen4}{rgb}{0.41,0.55,0.41}
\definecolor{DarkSeaGreen}{rgb}{0.56,0.74,0.56}
\definecolor{DarkSlateBlue}{rgb}{0.28,0.24,0.55}
\definecolor{DarkSlateGray1}{rgb}{0.59,1.00,1.00}
\definecolor{DarkSlateGray2}{rgb}{0.55,0.93,0.93}
\definecolor{DarkSlateGray3}{rgb}{0.47,0.80,0.80}
\definecolor{DarkSlateGray4}{rgb}{0.32,0.55,0.55}
\definecolor{DarkSlateGray}{rgb}{0.18,0.31,0.31}
\definecolor{DarkSlateGrey}{rgb}{0.18,0.31,0.31}
\definecolor{DarkTurquoise}{rgb}{0.00,0.81,0.82}
\definecolor{DarkViolet}{rgb}{0.58,0.00,0.83}
\definecolor{DeepPink1}{rgb}{1.00,0.08,0.58}
\definecolor{DeepPink2}{rgb}{0.93,0.07,0.54}
\definecolor{DeepPink3}{rgb}{0.80,0.06,0.46}
\definecolor{DeepPink4}{rgb}{0.55,0.04,0.31}
\definecolor{DeepPink}{rgb}{1.00,0.08,0.58}
\definecolor{DeepSkyBlue1}{rgb}{0.00,0.75,1.00}
\definecolor{DeepSkyBlue2}{rgb}{0.00,0.70,0.93}
\definecolor{DeepSkyBlue3}{rgb}{0.00,0.60,0.80}
\definecolor{DeepSkyBlue4}{rgb}{0.00,0.41,0.55}
\definecolor{DeepSkyBlue}{rgb}{0.00,0.75,1.00}
\definecolor{DimGray}{rgb}{0.41,0.41,0.41}
\definecolor{DimGrey}{rgb}{0.41,0.41,0.41}
\definecolor{DodgerBlue1}{rgb}{0.12,0.56,1.00}
\definecolor{DodgerBlue2}{rgb}{0.11,0.53,0.93}
\definecolor{DodgerBlue3}{rgb}{0.09,0.45,0.80}
\definecolor{DodgerBlue4}{rgb}{0.06,0.31,0.55}
\definecolor{DodgerBlue}{rgb}{0.12,0.56,1.00}
\definecolor{FloralWhite}{rgb}{1.00,0.98,0.94}
\definecolor{ForestGreen}{rgb}{0.13,0.55,0.13}
\definecolor{GhostWhite}{rgb}{0.97,0.97,1.00}
\definecolor{GreenYellow}{rgb}{0.68,1.00,0.18}
\definecolor{HotPink1}{rgb}{1.00,0.43,0.71}
\definecolor{HotPink2}{rgb}{0.93,0.42,0.65}
\definecolor{HotPink3}{rgb}{0.80,0.38,0.56}
\definecolor{HotPink4}{rgb}{0.55,0.23,0.38}
\definecolor{HotPink}{rgb}{1.00,0.41,0.71}
\definecolor{IndianRed1}{rgb}{1.00,0.42,0.42}
\definecolor{IndianRed2}{rgb}{0.93,0.39,0.39}
\definecolor{IndianRed3}{rgb}{0.80,0.33,0.33}
\definecolor{IndianRed4}{rgb}{0.55,0.23,0.23}
\definecolor{IndianRed}{rgb}{0.80,0.36,0.36}
\definecolor{LavenderBlush1}{rgb}{1.00,0.94,0.96}
\definecolor{LavenderBlush2}{rgb}{0.93,0.88,0.90}
\definecolor{LavenderBlush3}{rgb}{0.80,0.76,0.77}
\definecolor{LavenderBlush4}{rgb}{0.55,0.51,0.53}
\definecolor{LavenderBlush}{rgb}{1.00,0.94,0.96}
\definecolor{LawnGreen}{rgb}{0.49,0.99,0.00}
\definecolor{LemonChiffon1}{rgb}{1.00,0.98,0.80}
\definecolor{LemonChiffon2}{rgb}{0.93,0.91,0.75}
\definecolor{LemonChiffon3}{rgb}{0.80,0.79,0.65}
\definecolor{LemonChiffon4}{rgb}{0.55,0.54,0.44}
\definecolor{LemonChiffon}{rgb}{1.00,0.98,0.80}
\definecolor{LightBlue1}{rgb}{0.75,0.94,1.00}
\definecolor{LightBlue2}{rgb}{0.70,0.87,0.93}
\definecolor{LightBlue3}{rgb}{0.60,0.75,0.80}
\definecolor{LightBlue4}{rgb}{0.41,0.51,0.55}
\definecolor{LightBlue}{rgb}{0.68,0.85,0.90}
\definecolor{LightCoral}{rgb}{0.94,0.50,0.50}
\definecolor{LightCyan1}{rgb}{0.88,1.00,1.00}
\definecolor{LightCyan2}{rgb}{0.82,0.93,0.93}
\definecolor{LightCyan3}{rgb}{0.71,0.80,0.80}
\definecolor{LightCyan4}{rgb}{0.48,0.55,0.55}
\definecolor{LightCyan}{rgb}{0.88,1.00,1.00}
\definecolor{LightGoldenrod1}{rgb}{1.00,0.93,0.55}
\definecolor{LightGoldenrod2}{rgb}{0.93,0.86,0.51}
\definecolor{LightGoldenrod3}{rgb}{0.80,0.75,0.44}
\definecolor{LightGoldenrod4}{rgb}{0.55,0.51,0.30}
\definecolor{LightGoldenrodYellow}{rgb}{0.98,0.98,0.82}
\definecolor{LightGoldenrod}{rgb}{0.93,0.87,0.51}
\definecolor{LightGray}{rgb}{0.83,0.83,0.83}
\definecolor{LightGreen}{rgb}{0.56,0.93,0.56}
\definecolor{LightGrey}{rgb}{0.83,0.83,0.83}
\definecolor{LightPink1}{rgb}{1.00,0.68,0.73}
\definecolor{LightPink2}{rgb}{0.93,0.64,0.68}
\definecolor{LightPink3}{rgb}{0.80,0.55,0.58}
\definecolor{LightPink4}{rgb}{0.55,0.37,0.40}
\definecolor{LightPink}{rgb}{1.00,0.71,0.76}
\definecolor{LightSalmon1}{rgb}{1.00,0.63,0.48}
\definecolor{LightSalmon2}{rgb}{0.93,0.58,0.45}
\definecolor{LightSalmon3}{rgb}{0.80,0.51,0.38}
\definecolor{LightSalmon4}{rgb}{0.55,0.34,0.26}
\definecolor{LightSalmon}{rgb}{1.00,0.63,0.48}
\definecolor{LightSeaGreen}{rgb}{0.13,0.70,0.67}
\definecolor{LightSkyBlue1}{rgb}{0.69,0.89,1.00}
\definecolor{LightSkyBlue2}{rgb}{0.64,0.83,0.93}
\definecolor{LightSkyBlue3}{rgb}{0.55,0.71,0.80}
\definecolor{LightSkyBlue4}{rgb}{0.38,0.48,0.55}
\definecolor{LightSkyBlue}{rgb}{0.53,0.81,0.98}
\definecolor{LightSlateBlue}{rgb}{0.52,0.44,1.00}
\definecolor{LightSlateGray}{rgb}{0.47,0.53,0.60}
\definecolor{LightSlateGrey}{rgb}{0.47,0.53,0.60}
\definecolor{LightSteelBlue1}{rgb}{0.79,0.88,1.00}
\definecolor{LightSteelBlue2}{rgb}{0.74,0.82,0.93}
\definecolor{LightSteelBlue3}{rgb}{0.64,0.71,0.80}
\definecolor{LightSteelBlue4}{rgb}{0.43,0.48,0.55}
\definecolor{LightSteelBlue}{rgb}{0.69,0.77,0.87}
\definecolor{LightYellow1}{rgb}{1.00,1.00,0.88}
\definecolor{LightYellow2}{rgb}{0.93,0.93,0.82}
\definecolor{LightYellow3}{rgb}{0.80,0.80,0.71}
\definecolor{LightYellow4}{rgb}{0.55,0.55,0.48}
\definecolor{LightYellow}{rgb}{1.00,1.00,0.88}
\definecolor{LimeGreen}{rgb}{0.20,0.80,0.20}
\definecolor{MediumAquamarine}{rgb}{0.40,0.80,0.67}
\definecolor{MediumBlue}{rgb}{0.00,0.00,0.80}
\definecolor{MediumOrchid1}{rgb}{0.88,0.40,1.00}
\definecolor{MediumOrchid2}{rgb}{0.82,0.37,0.93}
\definecolor{MediumOrchid3}{rgb}{0.71,0.32,0.80}
\definecolor{MediumOrchid4}{rgb}{0.48,0.22,0.55}
\definecolor{MediumOrchid}{rgb}{0.73,0.33,0.83}
\definecolor{MediumPurple1}{rgb}{0.67,0.51,1.00}
\definecolor{MediumPurple2}{rgb}{0.62,0.47,0.93}
\definecolor{MediumPurple3}{rgb}{0.54,0.41,0.80}
\definecolor{MediumPurple4}{rgb}{0.36,0.28,0.55}
\definecolor{MediumPurple}{rgb}{0.58,0.44,0.86}
\definecolor{MediumSeaGreen}{rgb}{0.24,0.70,0.44}
\definecolor{MediumSlateBlue}{rgb}{0.48,0.41,0.93}
\definecolor{MediumSpringGreen}{rgb}{0.00,0.98,0.60}
\definecolor{MediumTurquoise}{rgb}{0.28,0.82,0.80}
\definecolor{MediumVioletRed}{rgb}{0.78,0.08,0.52}
\definecolor{MidnightBlue}{rgb}{0.10,0.10,0.44}
\definecolor{MintCream}{rgb}{0.96,1.00,0.98}
\definecolor{MistyRose1}{rgb}{1.00,0.89,0.88}
\definecolor{MistyRose2}{rgb}{0.93,0.84,0.82}
\definecolor{MistyRose3}{rgb}{0.80,0.72,0.71}
\definecolor{MistyRose4}{rgb}{0.55,0.49,0.48}
\definecolor{MistyRose}{rgb}{1.00,0.89,0.88}
\definecolor{NavajoWhite1}{rgb}{1.00,0.87,0.68}
\definecolor{NavajoWhite2}{rgb}{0.93,0.81,0.63}
\definecolor{NavajoWhite3}{rgb}{0.80,0.70,0.55}
\definecolor{NavajoWhite4}{rgb}{0.55,0.47,0.37}
\definecolor{NavajoWhite}{rgb}{1.00,0.87,0.68}
\definecolor{NavyBlue}{rgb}{0.00,0.00,0.50}
\definecolor{OldLace}{rgb}{0.99,0.96,0.90}
\definecolor{OliveDrab1}{rgb}{0.75,1.00,0.24}
\definecolor{OliveDrab2}{rgb}{0.70,0.93,0.23}
\definecolor{OliveDrab3}{rgb}{0.60,0.80,0.20}
\definecolor{OliveDrab4}{rgb}{0.41,0.55,0.13}
\definecolor{OliveDrab}{rgb}{0.42,0.56,0.14}
\definecolor{OrangeRed1}{rgb}{1.00,0.27,0.00}
\definecolor{OrangeRed2}{rgb}{0.93,0.25,0.00}
\definecolor{OrangeRed3}{rgb}{0.80,0.22,0.00}
\definecolor{OrangeRed4}{rgb}{0.55,0.15,0.00}
\definecolor{OrangeRed}{rgb}{1.00,0.27,0.00}
\definecolor{PaleGoldenrod}{rgb}{0.93,0.91,0.67}
\definecolor{PaleGreen1}{rgb}{0.60,1.00,0.60}
\definecolor{PaleGreen2}{rgb}{0.56,0.93,0.56}
\definecolor{PaleGreen3}{rgb}{0.49,0.80,0.49}
\definecolor{PaleGreen4}{rgb}{0.33,0.55,0.33}
\definecolor{PaleGreen}{rgb}{0.60,0.98,0.60}
\definecolor{PaleTurquoise1}{rgb}{0.73,1.00,1.00}
\definecolor{PaleTurquoise2}{rgb}{0.68,0.93,0.93}
\definecolor{PaleTurquoise3}{rgb}{0.59,0.80,0.80}
\definecolor{PaleTurquoise4}{rgb}{0.40,0.55,0.55}
\definecolor{PaleTurquoise}{rgb}{0.69,0.93,0.93}
\definecolor{PaleVioletRed1}{rgb}{1.00,0.51,0.67}
\definecolor{PaleVioletRed2}{rgb}{0.93,0.47,0.62}
\definecolor{PaleVioletRed3}{rgb}{0.80,0.41,0.54}
\definecolor{PaleVioletRed4}{rgb}{0.55,0.28,0.36}
\definecolor{PaleVioletRed}{rgb}{0.86,0.44,0.58}
\definecolor{PapayaWhip}{rgb}{1.00,0.94,0.84}
\definecolor{PeachPuff1}{rgb}{1.00,0.85,0.73}
\definecolor{PeachPuff2}{rgb}{0.93,0.80,0.68}
\definecolor{PeachPuff3}{rgb}{0.80,0.69,0.58}
\definecolor{PeachPuff4}{rgb}{0.55,0.47,0.40}
\definecolor{PeachPuff}{rgb}{1.00,0.85,0.73}
\definecolor{PowderBlue}{rgb}{0.69,0.88,0.90}
\definecolor{RosyBrown1}{rgb}{1.00,0.76,0.76}
\definecolor{RosyBrown2}{rgb}{0.93,0.71,0.71}
\definecolor{RosyBrown3}{rgb}{0.80,0.61,0.61}
\definecolor{RosyBrown4}{rgb}{0.55,0.41,0.41}
\definecolor{RosyBrown}{rgb}{0.74,0.56,0.56}
\definecolor{RoyalBlue1}{rgb}{0.28,0.46,1.00}
\definecolor{RoyalBlue2}{rgb}{0.26,0.43,0.93}
\definecolor{RoyalBlue3}{rgb}{0.23,0.37,0.80}
\definecolor{RoyalBlue4}{rgb}{0.15,0.25,0.55}
\definecolor{RoyalBlue}{rgb}{0.25,0.41,0.88}
\definecolor{SaddleBrown}{rgb}{0.55,0.27,0.07}
\definecolor{SandyBrown}{rgb}{0.96,0.64,0.38}
\definecolor{SeaGreen1}{rgb}{0.33,1.00,0.62}
\definecolor{SeaGreen2}{rgb}{0.31,0.93,0.58}
\definecolor{SeaGreen3}{rgb}{0.26,0.80,0.50}
\definecolor{SeaGreen4}{rgb}{0.18,0.55,0.34}
\definecolor{SeaGreen}{rgb}{0.18,0.55,0.34}
\definecolor{SkyBlue1}{rgb}{0.53,0.81,1.00}
\definecolor{SkyBlue2}{rgb}{0.49,0.75,0.93}
\definecolor{SkyBlue3}{rgb}{0.42,0.65,0.80}
\definecolor{SkyBlue4}{rgb}{0.29,0.44,0.55}
\definecolor{SkyBlue}{rgb}{0.53,0.81,0.92}
\definecolor{SlateBlue1}{rgb}{0.51,0.44,1.00}
\definecolor{SlateBlue2}{rgb}{0.48,0.40,0.93}
\definecolor{SlateBlue3}{rgb}{0.41,0.35,0.80}
\definecolor{SlateBlue4}{rgb}{0.28,0.24,0.55}
\definecolor{SlateBlue}{rgb}{0.42,0.35,0.80}
\definecolor{SlateGray1}{rgb}{0.78,0.89,1.00}
\definecolor{SlateGray2}{rgb}{0.73,0.83,0.93}
\definecolor{SlateGray3}{rgb}{0.62,0.71,0.80}
\definecolor{SlateGray4}{rgb}{0.42,0.48,0.55}
\definecolor{SlateGray}{rgb}{0.44,0.50,0.56}
\definecolor{SlateGrey}{rgb}{0.44,0.50,0.56}
\definecolor{SpringGreen1}{rgb}{0.00,1.00,0.50}
\definecolor{SpringGreen2}{rgb}{0.00,0.93,0.46}
\definecolor{SpringGreen3}{rgb}{0.00,0.80,0.40}
\definecolor{SpringGreen4}{rgb}{0.00,0.55,0.27}
\definecolor{SpringGreen}{rgb}{0.00,1.00,0.50}
\definecolor{SteelBlue1}{rgb}{0.39,0.72,1.00}
\definecolor{SteelBlue2}{rgb}{0.36,0.67,0.93}
\definecolor{SteelBlue3}{rgb}{0.31,0.58,0.80}
\definecolor{SteelBlue4}{rgb}{0.21,0.39,0.55}
\definecolor{SteelBlue}{rgb}{0.27,0.51,0.71}
\definecolor{VioletRed1}{rgb}{1.00,0.24,0.59}
\definecolor{VioletRed2}{rgb}{0.93,0.23,0.55}
\definecolor{VioletRed3}{rgb}{0.80,0.20,0.47}
\definecolor{VioletRed4}{rgb}{0.55,0.13,0.32}
\definecolor{VioletRed}{rgb}{0.82,0.13,0.56}
\definecolor{WhiteSmoke}{rgb}{0.96,0.96,0.96}
\definecolor{YellowGreen}{rgb}{0.60,0.80,0.20}
\definecolor{aliceblue}{rgb}{0.94,0.97,1.00}
\definecolor{antiquewhite}{rgb}{0.98,0.92,0.84}
\definecolor{aquamarine1}{rgb}{0.50,1.00,0.83}
\definecolor{aquamarine2}{rgb}{0.46,0.93,0.78}
\definecolor{aquamarine3}{rgb}{0.40,0.80,0.67}
\definecolor{aquamarine4}{rgb}{0.27,0.55,0.45}
\definecolor{aquamarine}{rgb}{0.50,1.00,0.83}
\definecolor{azure1}{rgb}{0.94,1.00,1.00}
\definecolor{azure2}{rgb}{0.88,0.93,0.93}
\definecolor{azure3}{rgb}{0.76,0.80,0.80}
\definecolor{azure4}{rgb}{0.51,0.55,0.55}
\definecolor{azure}{rgb}{0.94,1.00,1.00}
\definecolor{beige}{rgb}{0.96,0.96,0.86}
\definecolor{bisque1}{rgb}{1.00,0.89,0.77}
\definecolor{bisque2}{rgb}{0.93,0.84,0.72}
\definecolor{bisque3}{rgb}{0.80,0.72,0.62}
\definecolor{bisque4}{rgb}{0.55,0.49,0.42}
\definecolor{bisque}{rgb}{1.00,0.89,0.77}
\definecolor{black}{rgb}{0.00,0.00,0.00}
\definecolor{blanchedalmond}{rgb}{1.00,0.92,0.80}
\definecolor{blue1}{rgb}{0.00,0.00,1.00}
\definecolor{blue2}{rgb}{0.00,0.00,0.93}
\definecolor{blue3}{rgb}{0.00,0.00,0.80}
\definecolor{blue4}{rgb}{0.00,0.00,0.55}
\definecolor{blueviolet}{rgb}{0.54,0.17,0.89}
\definecolor{blue}{rgb}{0.00,0.00,1.00}
\definecolor{brown1}{rgb}{1.00,0.25,0.25}
\definecolor{brown2}{rgb}{0.93,0.23,0.23}
\definecolor{brown3}{rgb}{0.80,0.20,0.20}
\definecolor{brown4}{rgb}{0.55,0.14,0.14}
\definecolor{brown}{rgb}{0.65,0.16,0.16}
\definecolor{burlywood1}{rgb}{1.00,0.83,0.61}
\definecolor{burlywood2}{rgb}{0.93,0.77,0.57}
\definecolor{burlywood3}{rgb}{0.80,0.67,0.49}
\definecolor{burlywood4}{rgb}{0.55,0.45,0.33}
\definecolor{burlywood}{rgb}{0.87,0.72,0.53}
\definecolor{cadetblue}{rgb}{0.37,0.62,0.63}
\definecolor{chartreuse1}{rgb}{0.50,1.00,0.00}
\definecolor{chartreuse2}{rgb}{0.46,0.93,0.00}
\definecolor{chartreuse3}{rgb}{0.40,0.80,0.00}
\definecolor{chartreuse4}{rgb}{0.27,0.55,0.00}
\definecolor{chartreuse}{rgb}{0.50,1.00,0.00}
\definecolor{chocolate1}{rgb}{1.00,0.50,0.14}
\definecolor{chocolate2}{rgb}{0.93,0.46,0.13}
\definecolor{chocolate3}{rgb}{0.80,0.40,0.11}
\definecolor{chocolate4}{rgb}{0.55,0.27,0.07}
\definecolor{chocolate}{rgb}{0.82,0.41,0.12}
\definecolor{coral1}{rgb}{1.00,0.45,0.34}
\definecolor{coral2}{rgb}{0.93,0.42,0.31}
\definecolor{coral3}{rgb}{0.80,0.36,0.27}
\definecolor{coral4}{rgb}{0.55,0.24,0.18}
\definecolor{coral}{rgb}{1.00,0.50,0.31}
\definecolor{cornflowerblue}{rgb}{0.39,0.58,0.93}
\definecolor{cornsilk1}{rgb}{1.00,0.97,0.86}
\definecolor{cornsilk2}{rgb}{0.93,0.91,0.80}
\definecolor{cornsilk3}{rgb}{0.80,0.78,0.69}
\definecolor{cornsilk4}{rgb}{0.55,0.53,0.47}
\definecolor{cornsilk}{rgb}{1.00,0.97,0.86}
\definecolor{cyan1}{rgb}{0.00,1.00,1.00}
\definecolor{cyan2}{rgb}{0.00,0.93,0.93}
\definecolor{cyan3}{rgb}{0.00,0.80,0.80}
\definecolor{cyan4}{rgb}{0.00,0.55,0.55}
\definecolor{cyan}{rgb}{0.00,1.00,1.00}
\definecolor{darkblue}{rgb}{0.00,0.00,0.55}
\definecolor{darkcyan}{rgb}{0.00,0.55,0.55}
\definecolor{darkgoldenrod}{rgb}{0.72,0.53,0.04}
\definecolor{darkgray}{rgb}{0.66,0.66,0.66}
\definecolor{darkgreen}{rgb}{0.00,0.39,0.00}
\definecolor{darkgrey}{rgb}{0.66,0.66,0.66}
\definecolor{darkkhaki}{rgb}{0.74,0.72,0.42}
\definecolor{darkmagenta}{rgb}{0.55,0.00,0.55}
\definecolor{darkolive}{rgb}{0.33,0.42,0.18}
\definecolor{darkorange}{rgb}{1.00,0.55,0.00}
\definecolor{darkorchid}{rgb}{0.60,0.20,0.80}
\definecolor{darkred}{rgb}{0.55,0.00,0.00}
\definecolor{darksalmon}{rgb}{0.91,0.59,0.48}
\definecolor{darksea}{rgb}{0.56,0.74,0.56}
\definecolor{darkslate}{rgb}{0.18,0.31,0.31}
\definecolor{darkslate}{rgb}{0.18,0.31,0.31}
\definecolor{darkslate}{rgb}{0.28,0.24,0.55}
\definecolor{darkturquoise}{rgb}{0.00,0.81,0.82}
\definecolor{darkviolet}{rgb}{0.58,0.00,0.83}
\definecolor{deeppink}{rgb}{1.00,0.08,0.58}
\definecolor{deepsky}{rgb}{0.00,0.75,1.00}
\definecolor{dimgray}{rgb}{0.41,0.41,0.41}
\definecolor{dimgrey}{rgb}{0.41,0.41,0.41}
\definecolor{dodgerblue}{rgb}{0.12,0.56,1.00}
\definecolor{firebrick1}{rgb}{1.00,0.19,0.19}
\definecolor{firebrick2}{rgb}{0.93,0.17,0.17}
\definecolor{firebrick3}{rgb}{0.80,0.15,0.15}
\definecolor{firebrick4}{rgb}{0.55,0.10,0.10}
\definecolor{firebrick}{rgb}{0.70,0.13,0.13}
\definecolor{floralwhite}{rgb}{1.00,0.98,0.94}
\definecolor{forestgreen}{rgb}{0.13,0.55,0.13}
\definecolor{gainsboro}{rgb}{0.86,0.86,0.86}
\definecolor{ghostwhite}{rgb}{0.97,0.97,1.00}
\definecolor{gold1}{rgb}{1.00,0.84,0.00}
\definecolor{gold2}{rgb}{0.93,0.79,0.00}
\definecolor{gold3}{rgb}{0.80,0.68,0.00}
\definecolor{gold4}{rgb}{0.55,0.46,0.00}
\definecolor{goldenrod1}{rgb}{1.00,0.76,0.15}
\definecolor{goldenrod2}{rgb}{0.93,0.71,0.13}
\definecolor{goldenrod3}{rgb}{0.80,0.61,0.11}
\definecolor{goldenrod4}{rgb}{0.55,0.41,0.08}
\definecolor{goldenrod}{rgb}{0.85,0.65,0.13}
\definecolor{gold}{rgb}{1.00,0.84,0.00}
\definecolor{gray0}{rgb}{0.00,0.00,0.00}
\definecolor{gray100}{rgb}{1.00,1.00,1.00}
\definecolor{gray10}{rgb}{0.10,0.10,0.10}
\definecolor{gray11}{rgb}{0.11,0.11,0.11}
\definecolor{gray12}{rgb}{0.12,0.12,0.12}
\definecolor{gray13}{rgb}{0.13,0.13,0.13}
\definecolor{gray14}{rgb}{0.14,0.14,0.14}
\definecolor{gray15}{rgb}{0.15,0.15,0.15}
\definecolor{gray16}{rgb}{0.16,0.16,0.16}
\definecolor{gray17}{rgb}{0.17,0.17,0.17}
\definecolor{gray18}{rgb}{0.18,0.18,0.18}
\definecolor{gray19}{rgb}{0.19,0.19,0.19}
\definecolor{gray1}{rgb}{0.01,0.01,0.01}
\definecolor{gray20}{rgb}{0.20,0.20,0.20}
\definecolor{gray21}{rgb}{0.21,0.21,0.21}
\definecolor{gray22}{rgb}{0.22,0.22,0.22}
\definecolor{gray23}{rgb}{0.23,0.23,0.23}
\definecolor{gray24}{rgb}{0.24,0.24,0.24}
\definecolor{gray25}{rgb}{0.25,0.25,0.25}
\definecolor{gray26}{rgb}{0.26,0.26,0.26}
\definecolor{gray27}{rgb}{0.27,0.27,0.27}
\definecolor{gray28}{rgb}{0.28,0.28,0.28}
\definecolor{gray29}{rgb}{0.29,0.29,0.29}
\definecolor{gray2}{rgb}{0.02,0.02,0.02}
\definecolor{gray30}{rgb}{0.30,0.30,0.30}
\definecolor{gray31}{rgb}{0.31,0.31,0.31}
\definecolor{gray32}{rgb}{0.32,0.32,0.32}
\definecolor{gray33}{rgb}{0.33,0.33,0.33}
\definecolor{gray34}{rgb}{0.34,0.34,0.34}
\definecolor{gray35}{rgb}{0.35,0.35,0.35}
\definecolor{gray36}{rgb}{0.36,0.36,0.36}
\definecolor{gray37}{rgb}{0.37,0.37,0.37}
\definecolor{gray38}{rgb}{0.38,0.38,0.38}
\definecolor{gray39}{rgb}{0.39,0.39,0.39}
\definecolor{gray3}{rgb}{0.03,0.03,0.03}
\definecolor{gray40}{rgb}{0.40,0.40,0.40}
\definecolor{gray41}{rgb}{0.41,0.41,0.41}
\definecolor{gray42}{rgb}{0.42,0.42,0.42}
\definecolor{gray43}{rgb}{0.43,0.43,0.43}
\definecolor{gray44}{rgb}{0.44,0.44,0.44}
\definecolor{gray45}{rgb}{0.45,0.45,0.45}
\definecolor{gray46}{rgb}{0.46,0.46,0.46}
\definecolor{gray47}{rgb}{0.47,0.47,0.47}
\definecolor{gray48}{rgb}{0.48,0.48,0.48}
\definecolor{gray49}{rgb}{0.49,0.49,0.49}
\definecolor{gray4}{rgb}{0.04,0.04,0.04}
\definecolor{gray50}{rgb}{0.50,0.50,0.50}
\definecolor{gray51}{rgb}{0.51,0.51,0.51}
\definecolor{gray52}{rgb}{0.52,0.52,0.52}
\definecolor{gray53}{rgb}{0.53,0.53,0.53}
\definecolor{gray54}{rgb}{0.54,0.54,0.54}
\definecolor{gray55}{rgb}{0.55,0.55,0.55}
\definecolor{gray56}{rgb}{0.56,0.56,0.56}
\definecolor{gray57}{rgb}{0.57,0.57,0.57}
\definecolor{gray58}{rgb}{0.58,0.58,0.58}
\definecolor{gray59}{rgb}{0.59,0.59,0.59}
\definecolor{gray5}{rgb}{0.05,0.05,0.05}
\definecolor{gray60}{rgb}{0.60,0.60,0.60}
\definecolor{gray61}{rgb}{0.61,0.61,0.61}
\definecolor{gray62}{rgb}{0.62,0.62,0.62}
\definecolor{gray63}{rgb}{0.63,0.63,0.63}
\definecolor{gray64}{rgb}{0.64,0.64,0.64}
\definecolor{gray65}{rgb}{0.65,0.65,0.65}
\definecolor{gray66}{rgb}{0.66,0.66,0.66}
\definecolor{gray67}{rgb}{0.67,0.67,0.67}
\definecolor{gray68}{rgb}{0.68,0.68,0.68}
\definecolor{gray69}{rgb}{0.69,0.69,0.69}
\definecolor{gray6}{rgb}{0.06,0.06,0.06}
\definecolor{gray70}{rgb}{0.70,0.70,0.70}
\definecolor{gray71}{rgb}{0.71,0.71,0.71}
\definecolor{gray72}{rgb}{0.72,0.72,0.72}
\definecolor{gray73}{rgb}{0.73,0.73,0.73}
\definecolor{gray74}{rgb}{0.74,0.74,0.74}
\definecolor{gray75}{rgb}{0.75,0.75,0.75}
\definecolor{gray76}{rgb}{0.76,0.76,0.76}
\definecolor{gray77}{rgb}{0.77,0.77,0.77}
\definecolor{gray78}{rgb}{0.78,0.78,0.78}
\definecolor{gray79}{rgb}{0.79,0.79,0.79}
\definecolor{gray7}{rgb}{0.07,0.07,0.07}
\definecolor{gray80}{rgb}{0.80,0.80,0.80}
\definecolor{gray81}{rgb}{0.81,0.81,0.81}
\definecolor{gray82}{rgb}{0.82,0.82,0.82}
\definecolor{gray83}{rgb}{0.83,0.83,0.83}
\definecolor{gray84}{rgb}{0.84,0.84,0.84}
\definecolor{gray85}{rgb}{0.85,0.85,0.85}
\definecolor{gray86}{rgb}{0.86,0.86,0.86}
\definecolor{gray87}{rgb}{0.87,0.87,0.87}
\definecolor{gray88}{rgb}{0.88,0.88,0.88}
\definecolor{gray89}{rgb}{0.89,0.89,0.89}
\definecolor{gray8}{rgb}{0.08,0.08,0.08}
\definecolor{gray90}{rgb}{0.90,0.90,0.90}
\definecolor{gray91}{rgb}{0.91,0.91,0.91}
\definecolor{gray92}{rgb}{0.92,0.92,0.92}
\definecolor{gray93}{rgb}{0.93,0.93,0.93}
\definecolor{gray94}{rgb}{0.94,0.94,0.94}
\definecolor{gray95}{rgb}{0.95,0.95,0.95}
\definecolor{gray96}{rgb}{0.96,0.96,0.96}
\definecolor{gray97}{rgb}{0.97,0.97,0.97}
\definecolor{gray98}{rgb}{0.98,0.98,0.98}
\definecolor{gray99}{rgb}{0.99,0.99,0.99}
\definecolor{gray9}{rgb}{0.09,0.09,0.09}
\definecolor{gray}{rgb}{0.75,0.75,0.75}
\definecolor{green1}{rgb}{0.00,1.00,0.00}
\definecolor{green2}{rgb}{0.00,0.93,0.00}
\definecolor{green3}{rgb}{0.00,0.80,0.00}
\definecolor{green4}{rgb}{0.00,0.55,0.00}
\definecolor{greenyellow}{rgb}{0.68,1.00,0.18}
\definecolor{green}{rgb}{0.00,1.00,0.00}
\definecolor{grey0}{rgb}{0.00,0.00,0.00}
\definecolor{grey100}{rgb}{1.00,1.00,1.00}
\definecolor{grey10}{rgb}{0.10,0.10,0.10}
\definecolor{grey11}{rgb}{0.11,0.11,0.11}
\definecolor{grey12}{rgb}{0.12,0.12,0.12}
\definecolor{grey13}{rgb}{0.13,0.13,0.13}
\definecolor{grey14}{rgb}{0.14,0.14,0.14}
\definecolor{grey15}{rgb}{0.15,0.15,0.15}
\definecolor{grey16}{rgb}{0.16,0.16,0.16}
\definecolor{grey17}{rgb}{0.17,0.17,0.17}
\definecolor{grey18}{rgb}{0.18,0.18,0.18}
\definecolor{grey19}{rgb}{0.19,0.19,0.19}
\definecolor{grey1}{rgb}{0.01,0.01,0.01}
\definecolor{grey20}{rgb}{0.20,0.20,0.20}
\definecolor{grey21}{rgb}{0.21,0.21,0.21}
\definecolor{grey22}{rgb}{0.22,0.22,0.22}
\definecolor{grey23}{rgb}{0.23,0.23,0.23}
\definecolor{grey24}{rgb}{0.24,0.24,0.24}
\definecolor{grey25}{rgb}{0.25,0.25,0.25}
\definecolor{grey26}{rgb}{0.26,0.26,0.26}
\definecolor{grey27}{rgb}{0.27,0.27,0.27}
\definecolor{grey28}{rgb}{0.28,0.28,0.28}
\definecolor{grey29}{rgb}{0.29,0.29,0.29}
\definecolor{grey2}{rgb}{0.02,0.02,0.02}
\definecolor{grey30}{rgb}{0.30,0.30,0.30}
\definecolor{grey31}{rgb}{0.31,0.31,0.31}
\definecolor{grey32}{rgb}{0.32,0.32,0.32}
\definecolor{grey33}{rgb}{0.33,0.33,0.33}
\definecolor{grey34}{rgb}{0.34,0.34,0.34}
\definecolor{grey35}{rgb}{0.35,0.35,0.35}
\definecolor{grey36}{rgb}{0.36,0.36,0.36}
\definecolor{grey37}{rgb}{0.37,0.37,0.37}
\definecolor{grey38}{rgb}{0.38,0.38,0.38}
\definecolor{grey39}{rgb}{0.39,0.39,0.39}
\definecolor{grey3}{rgb}{0.03,0.03,0.03}
\definecolor{grey40}{rgb}{0.40,0.40,0.40}
\definecolor{grey41}{rgb}{0.41,0.41,0.41}
\definecolor{grey42}{rgb}{0.42,0.42,0.42}
\definecolor{grey43}{rgb}{0.43,0.43,0.43}
\definecolor{grey44}{rgb}{0.44,0.44,0.44}
\definecolor{grey45}{rgb}{0.45,0.45,0.45}
\definecolor{grey46}{rgb}{0.46,0.46,0.46}
\definecolor{grey47}{rgb}{0.47,0.47,0.47}
\definecolor{grey48}{rgb}{0.48,0.48,0.48}
\definecolor{grey49}{rgb}{0.49,0.49,0.49}
\definecolor{grey4}{rgb}{0.04,0.04,0.04}
\definecolor{grey50}{rgb}{0.50,0.50,0.50}
\definecolor{grey51}{rgb}{0.51,0.51,0.51}
\definecolor{grey52}{rgb}{0.52,0.52,0.52}
\definecolor{grey53}{rgb}{0.53,0.53,0.53}
\definecolor{grey54}{rgb}{0.54,0.54,0.54}
\definecolor{grey55}{rgb}{0.55,0.55,0.55}
\definecolor{grey56}{rgb}{0.56,0.56,0.56}
\definecolor{grey57}{rgb}{0.57,0.57,0.57}
\definecolor{grey58}{rgb}{0.58,0.58,0.58}
\definecolor{grey59}{rgb}{0.59,0.59,0.59}
\definecolor{grey5}{rgb}{0.05,0.05,0.05}
\definecolor{grey60}{rgb}{0.60,0.60,0.60}
\definecolor{grey61}{rgb}{0.61,0.61,0.61}
\definecolor{grey62}{rgb}{0.62,0.62,0.62}
\definecolor{grey63}{rgb}{0.63,0.63,0.63}
\definecolor{grey64}{rgb}{0.64,0.64,0.64}
\definecolor{grey65}{rgb}{0.65,0.65,0.65}
\definecolor{grey66}{rgb}{0.66,0.66,0.66}
\definecolor{grey67}{rgb}{0.67,0.67,0.67}
\definecolor{grey68}{rgb}{0.68,0.68,0.68}
\definecolor{grey69}{rgb}{0.69,0.69,0.69}
\definecolor{grey6}{rgb}{0.06,0.06,0.06}
\definecolor{grey70}{rgb}{0.70,0.70,0.70}
\definecolor{grey71}{rgb}{0.71,0.71,0.71}
\definecolor{grey72}{rgb}{0.72,0.72,0.72}
\definecolor{grey73}{rgb}{0.73,0.73,0.73}
\definecolor{grey74}{rgb}{0.74,0.74,0.74}
\definecolor{grey75}{rgb}{0.75,0.75,0.75}
\definecolor{grey76}{rgb}{0.76,0.76,0.76}
\definecolor{grey77}{rgb}{0.77,0.77,0.77}
\definecolor{grey78}{rgb}{0.78,0.78,0.78}
\definecolor{grey79}{rgb}{0.79,0.79,0.79}
\definecolor{grey7}{rgb}{0.07,0.07,0.07}
\definecolor{grey80}{rgb}{0.80,0.80,0.80}
\definecolor{grey81}{rgb}{0.81,0.81,0.81}
\definecolor{grey82}{rgb}{0.82,0.82,0.82}
\definecolor{grey83}{rgb}{0.83,0.83,0.83}
\definecolor{grey84}{rgb}{0.84,0.84,0.84}
\definecolor{grey85}{rgb}{0.85,0.85,0.85}
\definecolor{grey86}{rgb}{0.86,0.86,0.86}
\definecolor{grey87}{rgb}{0.87,0.87,0.87}
\definecolor{grey88}{rgb}{0.88,0.88,0.88}
\definecolor{grey89}{rgb}{0.89,0.89,0.89}
\definecolor{grey8}{rgb}{0.08,0.08,0.08}
\definecolor{grey90}{rgb}{0.90,0.90,0.90}
\definecolor{grey91}{rgb}{0.91,0.91,0.91}
\definecolor{grey92}{rgb}{0.92,0.92,0.92}
\definecolor{grey93}{rgb}{0.93,0.93,0.93}
\definecolor{grey94}{rgb}{0.94,0.94,0.94}
\definecolor{grey95}{rgb}{0.95,0.95,0.95}
\definecolor{grey96}{rgb}{0.96,0.96,0.96}
\definecolor{grey97}{rgb}{0.97,0.97,0.97}
\definecolor{grey98}{rgb}{0.98,0.98,0.98}
\definecolor{grey99}{rgb}{0.99,0.99,0.99}
\definecolor{grey9}{rgb}{0.09,0.09,0.09}
\definecolor{grey}{rgb}{0.75,0.75,0.75}
\definecolor{honeydew1}{rgb}{0.94,1.00,0.94}
\definecolor{honeydew2}{rgb}{0.88,0.93,0.88}
\definecolor{honeydew3}{rgb}{0.76,0.80,0.76}
\definecolor{honeydew4}{rgb}{0.51,0.55,0.51}
\definecolor{honeydew}{rgb}{0.94,1.00,0.94}
\definecolor{hotpink}{rgb}{1.00,0.41,0.71}
\definecolor{indianred}{rgb}{0.80,0.36,0.36}
\definecolor{ivory1}{rgb}{1.00,1.00,0.94}
\definecolor{ivory2}{rgb}{0.93,0.93,0.88}
\definecolor{ivory3}{rgb}{0.80,0.80,0.76}
\definecolor{ivory4}{rgb}{0.55,0.55,0.51}
\definecolor{ivory}{rgb}{1.00,1.00,0.94}
\definecolor{khaki1}{rgb}{1.00,0.96,0.56}
\definecolor{khaki2}{rgb}{0.93,0.90,0.52}
\definecolor{khaki3}{rgb}{0.80,0.78,0.45}
\definecolor{khaki4}{rgb}{0.55,0.53,0.31}
\definecolor{khaki}{rgb}{0.94,0.90,0.55}
\definecolor{lavenderblush}{rgb}{1.00,0.94,0.96}
\definecolor{lavender}{rgb}{0.90,0.90,0.98}
\definecolor{lawngreen}{rgb}{0.49,0.99,0.00}
\definecolor{lemonchiffon}{rgb}{1.00,0.98,0.80}
\definecolor{lightblue}{rgb}{0.68,0.85,0.90}
\definecolor{lightcoral}{rgb}{0.94,0.50,0.50}
\definecolor{lightcyan}{rgb}{0.88,1.00,1.00}
\definecolor{lightgoldenrod}{rgb}{0.93,0.87,0.51}
\definecolor{lightgoldenrod}{rgb}{0.98,0.98,0.82}
\definecolor{lightgray}{rgb}{0.83,0.83,0.83}
\definecolor{lightgreen}{rgb}{0.56,0.93,0.56}
\definecolor{lightgrey}{rgb}{0.83,0.83,0.83}
\definecolor{lightpink}{rgb}{1.00,0.71,0.76}
\definecolor{lightsalmon}{rgb}{1.00,0.63,0.48}
\definecolor{lightsea}{rgb}{0.13,0.70,0.67}
\definecolor{lightsky}{rgb}{0.53,0.81,0.98}
\definecolor{lightslate}{rgb}{0.47,0.53,0.60}
\definecolor{lightslate}{rgb}{0.47,0.53,0.60}
\definecolor{lightslate}{rgb}{0.52,0.44,1.00}
\definecolor{lightsteel}{rgb}{0.69,0.77,0.87}
\definecolor{lightyellow}{rgb}{1.00,1.00,0.88}
\definecolor{limegreen}{rgb}{0.20,0.80,0.20}
\definecolor{linen}{rgb}{0.98,0.94,0.90}
\definecolor{magenta1}{rgb}{1.00,0.00,1.00}
\definecolor{magenta2}{rgb}{0.93,0.00,0.93}
\definecolor{magenta3}{rgb}{0.80,0.00,0.80}
\definecolor{magenta4}{rgb}{0.55,0.00,0.55}
\definecolor{magenta}{rgb}{1.00,0.00,1.00}
\definecolor{maroon1}{rgb}{1.00,0.20,0.70}
\definecolor{maroon2}{rgb}{0.93,0.19,0.65}
\definecolor{maroon3}{rgb}{0.80,0.16,0.56}
\definecolor{maroon4}{rgb}{0.55,0.11,0.38}
\definecolor{maroon}{rgb}{0.69,0.19,0.38}
\definecolor{mediumaquamarine}{rgb}{0.40,0.80,0.67}
\definecolor{mediumblue}{rgb}{0.00,0.00,0.80}
\definecolor{mediumorchid}{rgb}{0.73,0.33,0.83}
\definecolor{mediumpurple}{rgb}{0.58,0.44,0.86}
\definecolor{mediumsea}{rgb}{0.24,0.70,0.44}
\definecolor{mediumslate}{rgb}{0.48,0.41,0.93}
\definecolor{mediumspring}{rgb}{0.00,0.98,0.60}
\definecolor{mediumturquoise}{rgb}{0.28,0.82,0.80}
\definecolor{mediumviolet}{rgb}{0.78,0.08,0.52}
\definecolor{midnightblue}{rgb}{0.10,0.10,0.44}
\definecolor{mintcream}{rgb}{0.96,1.00,0.98}
\definecolor{mistyrose}{rgb}{1.00,0.89,0.88}
\definecolor{moccasin}{rgb}{1.00,0.89,0.71}
\definecolor{navajowhite}{rgb}{1.00,0.87,0.68}
\definecolor{navyblue}{rgb}{0.00,0.00,0.50}
\definecolor{navy}{rgb}{0.00,0.00,0.50}
\definecolor{oldlace}{rgb}{0.99,0.96,0.90}
\definecolor{olivedrab}{rgb}{0.42,0.56,0.14}
\definecolor{orange1}{rgb}{1.00,0.65,0.00}
\definecolor{orange2}{rgb}{0.93,0.60,0.00}
\definecolor{orange3}{rgb}{0.80,0.52,0.00}
\definecolor{orange4}{rgb}{0.55,0.35,0.00}
\definecolor{orangered}{rgb}{1.00,0.27,0.00}
\definecolor{orange}{rgb}{1.00,0.65,0.00}
\definecolor{orchid1}{rgb}{1.00,0.51,0.98}
\definecolor{orchid2}{rgb}{0.93,0.48,0.91}
\definecolor{orchid3}{rgb}{0.80,0.41,0.79}
\definecolor{orchid4}{rgb}{0.55,0.28,0.54}
\definecolor{orchid}{rgb}{0.85,0.44,0.84}
\definecolor{palegoldenrod}{rgb}{0.93,0.91,0.67}
\definecolor{palegreen}{rgb}{0.60,0.98,0.60}
\definecolor{paleturquoise}{rgb}{0.69,0.93,0.93}
\definecolor{paleviolet}{rgb}{0.86,0.44,0.58}
\definecolor{papayawhip}{rgb}{1.00,0.94,0.84}
\definecolor{peachpuff}{rgb}{1.00,0.85,0.73}
\definecolor{peru}{rgb}{0.80,0.52,0.25}
\definecolor{pink1}{rgb}{1.00,0.71,0.77}
\definecolor{pink2}{rgb}{0.93,0.66,0.72}
\definecolor{pink3}{rgb}{0.80,0.57,0.62}
\definecolor{pink4}{rgb}{0.55,0.39,0.42}
\definecolor{pink}{rgb}{1.00,0.75,0.80}
\definecolor{plum1}{rgb}{1.00,0.73,1.00}
\definecolor{plum2}{rgb}{0.93,0.68,0.93}
\definecolor{plum3}{rgb}{0.80,0.59,0.80}
\definecolor{plum4}{rgb}{0.55,0.40,0.55}
\definecolor{plum}{rgb}{0.87,0.63,0.87}
\definecolor{powderblue}{rgb}{0.69,0.88,0.90}
\definecolor{purple1}{rgb}{0.61,0.19,1.00}
\definecolor{purple2}{rgb}{0.57,0.17,0.93}
\definecolor{purple3}{rgb}{0.49,0.15,0.80}
\definecolor{purple4}{rgb}{0.33,0.10,0.55}
\definecolor{purple}{rgb}{0.63,0.13,0.94}
\definecolor{red1}{rgb}{1.00,0.00,0.00}
\definecolor{red2}{rgb}{0.93,0.00,0.00}
\definecolor{red3}{rgb}{0.80,0.00,0.00}
\definecolor{red4}{rgb}{0.55,0.00,0.00}
\definecolor{red}{rgb}{1.00,0.00,0.00}
\definecolor{rosybrown}{rgb}{0.74,0.56,0.56}
\definecolor{royalblue}{rgb}{0.25,0.41,0.88}
\definecolor{saddlebrown}{rgb}{0.55,0.27,0.07}
\definecolor{salmon1}{rgb}{1.00,0.55,0.41}
\definecolor{salmon2}{rgb}{0.93,0.51,0.38}
\definecolor{salmon3}{rgb}{0.80,0.44,0.33}
\definecolor{salmon4}{rgb}{0.55,0.30,0.22}
\definecolor{salmon}{rgb}{0.98,0.50,0.45}
\definecolor{sandybrown}{rgb}{0.96,0.64,0.38}
\definecolor{seagreen}{rgb}{0.18,0.55,0.34}
\definecolor{seashell1}{rgb}{1.00,0.96,0.93}
\definecolor{seashell2}{rgb}{0.93,0.90,0.87}
\definecolor{seashell3}{rgb}{0.80,0.77,0.75}
\definecolor{seashell4}{rgb}{0.55,0.53,0.51}
\definecolor{seashell}{rgb}{1.00,0.96,0.93}
\definecolor{sienna1}{rgb}{1.00,0.51,0.28}
\definecolor{sienna2}{rgb}{0.93,0.47,0.26}
\definecolor{sienna3}{rgb}{0.80,0.41,0.22}
\definecolor{sienna4}{rgb}{0.55,0.28,0.15}
\definecolor{sienna}{rgb}{0.63,0.32,0.18}
\definecolor{skyblue}{rgb}{0.53,0.81,0.92}
\definecolor{slateblue}{rgb}{0.42,0.35,0.80}
\definecolor{slategray}{rgb}{0.44,0.50,0.56}
\definecolor{slategrey}{rgb}{0.44,0.50,0.56}
\definecolor{snow1}{rgb}{1.00,0.98,0.98}
\definecolor{snow2}{rgb}{0.93,0.91,0.91}
\definecolor{snow3}{rgb}{0.80,0.79,0.79}
\definecolor{snow4}{rgb}{0.55,0.54,0.54}
\definecolor{snow}{rgb}{1.00,0.98,0.98}
\definecolor{springgreen}{rgb}{0.00,1.00,0.50}
\definecolor{steelblue}{rgb}{0.27,0.51,0.71}
\definecolor{tan1}{rgb}{1.00,0.65,0.31}
\definecolor{tan2}{rgb}{0.93,0.60,0.29}
\definecolor{tan3}{rgb}{0.80,0.52,0.25}
\definecolor{tan4}{rgb}{0.55,0.35,0.17}
\definecolor{tan}{rgb}{0.82,0.71,0.55}
\definecolor{thistle1}{rgb}{1.00,0.88,1.00}
\definecolor{thistle2}{rgb}{0.93,0.82,0.93}
\definecolor{thistle3}{rgb}{0.80,0.71,0.80}
\definecolor{thistle4}{rgb}{0.55,0.48,0.55}
\definecolor{thistle}{rgb}{0.85,0.75,0.85}
\definecolor{tomato1}{rgb}{1.00,0.39,0.28}
\definecolor{tomato2}{rgb}{0.93,0.36,0.26}
\definecolor{tomato3}{rgb}{0.80,0.31,0.22}
\definecolor{tomato4}{rgb}{0.55,0.21,0.15}
\definecolor{tomato}{rgb}{1.00,0.39,0.28}
\definecolor{turquoise1}{rgb}{0.00,0.96,1.00}
\definecolor{turquoise2}{rgb}{0.00,0.90,0.93}
\definecolor{turquoise3}{rgb}{0.00,0.77,0.80}
\definecolor{turquoise4}{rgb}{0.00,0.53,0.55}
\definecolor{turquoise}{rgb}{0.25,0.88,0.82}
\definecolor{violetred}{rgb}{0.82,0.13,0.56}
\definecolor{violet}{rgb}{0.93,0.51,0.93}
\definecolor{wheat1}{rgb}{1.00,0.91,0.73}
\definecolor{wheat2}{rgb}{0.93,0.85,0.68}
\definecolor{wheat3}{rgb}{0.80,0.73,0.59}
\definecolor{wheat4}{rgb}{0.55,0.49,0.40}
\definecolor{wheat}{rgb}{0.96,0.87,0.70}
\definecolor{whitesmoke}{rgb}{0.96,0.96,0.96}
\definecolor{white}{rgb}{1.00,1.00,1.00}
\definecolor{yellow1}{rgb}{1.00,1.00,0.00}
\definecolor{yellow2}{rgb}{0.93,0.93,0.00}
\definecolor{yellow3}{rgb}{0.80,0.80,0.00}
\definecolor{yellow4}{rgb}{0.55,0.55,0.00}
\definecolor{yellowgreen}{rgb}{0.60,0.80,0.20}
\definecolor{yellow}{rgb}{1.00,1.00,0.00}

%% file: defs.tex
\newcommand{\pluseq}{\mathrel{+}=}

\newcommand{\V}[1]{{\bf #1}}

\newcommand{\image}{\mathbf{c}}
\newcommand{\sdepth}{\mathbf{d}}
\newcommand{\sdepthmin}{\mathbf{d}_{min}}
\newcommand{\sdepthmax}{\mathbf{d}_{max}}
\newcommand{\softsplatweight}{\mathbf{w}}
\newcommand{\warpedcolor}{\overline{\mathbf{c}}}
\newcommand{\warpeddepth}{\overline{\mathbf{d}}}
\newcommand{\partoccfilteredcolor}{\widehat{\mathbf{c}}}
\newcommand{\fulloccfilteredcolor}{\mathbf{c}^\ast}
\newcommand{\finalcolor}{\mathbf{c}^\dagger}
\newcommand{\gtcolor}{\mathbf{c}^{ref}}

\newcommand{\occlusionmask}{\mathbf{m}}
\newcommand{\fullocclusionmask}{\widehat{\mathbf{m}}}

\newcommand{\metricssim}{\text{\em{ssim}}}

\newcommand{\psf}{\mathbf{k}}

\newcommand{\mfusion}{\text{\em{fusion}}}
\newcommand{\mfulldisocclusionfiltering}{\text{\em{full\_disocclusion\_filtering}}}

\newcommand{\disocclusionsize}{{\beta}}
\newcommand{\xoffset}{{\alpha}}
\newcommand{\ipd}{{\rho}}
\newcommand{\hmdthickness}{{\varphi}}
\newcommand{\fieldofinterest}{{\theta}}
\newcommand{\neardepth}{{d_n}}
\newcommand{\fardepth}{{d_f}}

\newcommand{\figref}[1]{Figure~\ref{#1}}
\newcommand{\secref}[1]{Section~\ref{#1}}

%% file: introduction.tex
\section{Introduction}
\label{sec:intro}

Virtual reality (VR) head-mounted displays (HMDs) provide nearly complete visual immersion, using a pair of near-eye displays to create wide-field-of-view, stereoscopic images. However, such immersion comes at the cost of visual isolation from the user's physical environment. For certain applications, a direct view of the nearby surroundings is necessary. To this end, augmented reality (AR) uses near-eye displays to support \emph{optical see-through}. Yet, modern AR displays achieve limited fields of view, unlike blocked-light VR; thus, \emph{video see-through} VR has been proposed as a solution, using \emph{passthrough} algorithms to transform imagery collected by outward-facing cameras to enable the user to see their surroundings.

In practice, VR passthrough systems do not directly pass through anything. Rather, they must accomplish the difficult task of reprojecting camera images to appear from the user’s perspective. This is often approximated, with Krajancich et al.~\shortcite{Krajancich2020gcr} showing the value of updating reconstructions to track the user’s moving pupils. While pupil-tracked passthrough may be the ultimate goal, state-of-the-art techniques, such as \emph{Passthrough+}~\cite{chaurasia2020passthrough+}, reproject camera data to the nominal position of the eyes, while accepting any artifacts resulting from the computational limits. 

Real-time view synthesis lies at the core of achieving compelling passthrough experiences. While this is a challenging problem itself~\cite{zhou2018stereo,mildenhall2019local,mildenhall2020nerf}, VR headsets present particularly daunting requirements that cannot be supported by many modern methods. Namely, commercial VR displays are stereoscopic, refresh at $72$-$144$ frames per second, and support wide fields of view ($>$$90^\circ$, horizontally). For VR passthrough, a typical scenario involves a user's manipulating near-field objects with their own hands and observing dynamic environments, resulting in large regions with missing data, due to disocclusions, and preventing offline reconstruction from prior observations.

\begin{figure}
	\centering
	\includegraphics[width=\columnwidth]{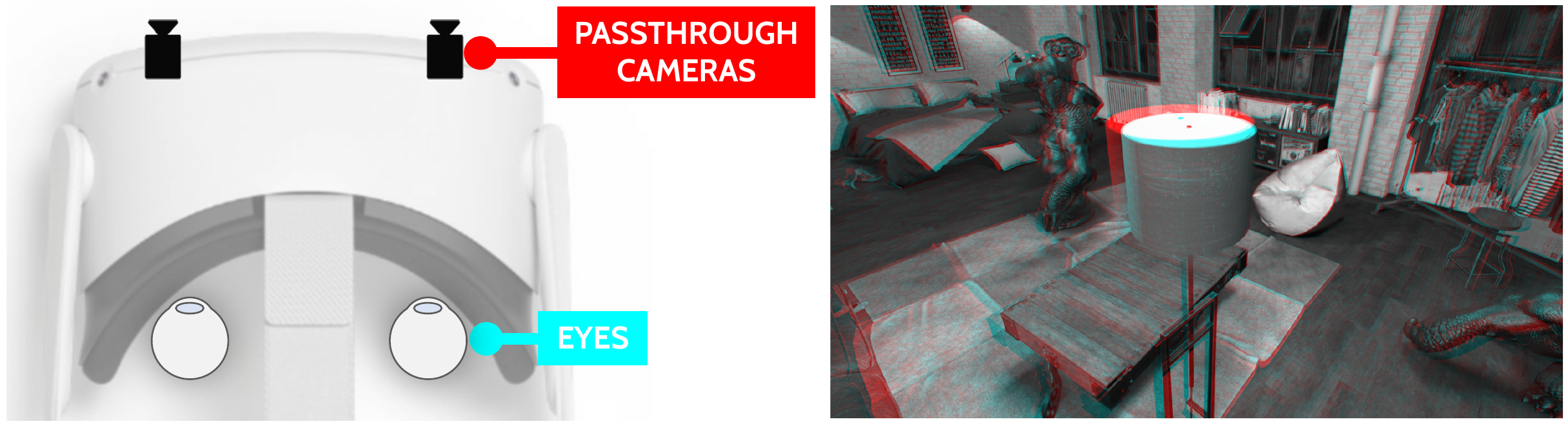}
	\caption{{\em{Left:}} Typically, VR passthrough cameras are located several centimeters away from the user's eyes. {\em{Right:}} Separation between the cameras and the eyes requires significant viewpoint changes to be resolved by the passthrough algorithm. Here the right-eye camera image (shown in red) is compared to what should be seen by the user's right eye (shown in cyan).}
	\label{fig:problem_setup}
\end{figure}

Given these algorithmic challenges, headset designers can assist passthrough by placing cameras as close to the user's eyes as possible, asking the algorithm to make only modest changes. Yet, as shown in~\figref{fig:problem_setup}, cameras cannot be co-located with the user's eyes. Thus, in this paper, we optimize the performance of a minimal passthrough architecture: placing a stereo pair of RGB cameras on the front of a VR headset (see~\figref{fig:teaser}). Consistent with Chaurasia et al.~\shortcite{chaurasia2020passthrough+}, we identify that such a configuration offers a practical trade-off between hardware and algorithmic complexity. However, unlike prior works, we consider the optimal placement of the cameras to work in concert with the passthrough algorithm, exploring how the baseline can be adjusted to mitigate reprojection artifacts.

We aim to find an efficient, high-quality method for real-time stereoscopic view synthesis from stereo inputs. Being the closest prior work, we seek to address the limitations of Chaurasia et al.~\shortcite{chaurasia2020passthrough+}, which applies a traditional 3D computer vision pipeline: for each frame, a sparse point cloud is reconstructed and processed to produce meshes for reprojection. In contrast, we introduce \emph{NeuralPassthrough} to leverage recent advances in deep learning, solving passthrough as an image-based neural rendering problem. Specifically, we jointly apply learned stereo depth estimation and image reconstruction networks to produce the eye-viewpoint images via an end-to-end approach, one that is tailored for today's desktop-connected VR headsets and their strict real-time requirements. {\newrevise{The source code and pretrained models will be available at https://research.facebook.com/publications/neural-passthrough/, pending institutional approval.}}
Our primary technical contributions include the following:
\begin{itemize}[topsep=0pt]
\item We build a VR headset with an adjustable stereo camera baseline, optimized for evaluating view synthesis methods.
\item We analyze the impact of camera placement on VR passthrough image quality; our analysis reveals that key disocclusions can be mitigated by adopting wider camera baselines than the user's interpupillary distance (IPD) — a design optimization that diverges from current consumer products.
\item We demonstrate the first learned view synthesis method tailored for real-time VR passthrough, suppressing key artifacts and achieving higher image quality than prior methods.
\end{itemize}

%% file: previous_work.tex
\section{Related Work}
\label{sec:related}

\subsection{Classical Methods for View Synthesis}
\label{sec:related:classical}

Early work on light field imaging addresses view synthesis as an interpolation problem, resampling captured rays to generate novel views~\cite{gortler1996lumigraph,levoy1996light}. However, by requiring a dense set of input views, such methods are not well suited for VR passthrough.
{\revise{View synthesis from sparse views gained much attention in graphics rendering~\cite{chen1993view,asw2}, computational photography~\cite{shade1998layered,zitnick2004high,hedman2017casual} and displays~\cite{chapiro2014optimizing,didyk2013joint}. Most of these methods require rendered depth as input, allow limited viewpoint changes, or run offline.}}

To our knowledge, \emph{Passthrough+}~\cite{chaurasia2020passthrough+} is the only prior work that directly addresses VR passthrough.
It is specifically tailored for mobile VR applications, but achieves limited image quality. In contrast, our method focuses on desktop-connected VR headsets, leveraging more compute resources while continuing to be strictly constrained by the same real-time requirements.

\subsection{Learning-based Methods for View Synthesis}
\label{sec:related:learning}

Significant recent progress has been made in learning-based view synthesis, which can be grouped according to the inputs: {\revise{single views~\cite{kopf2020one,shih3d,wiles2020synsin}}}; stereo views~\cite{zhou2018stereo}; multiple views~\cite{kalantari2016learning,flynn2019deepview,mildenhall2019local,mildenhall2020nerf}; multiview videos~\cite{broxton2020immersive}; {\revise{and RGB-D captures~\cite{martin2018lookingood,aliev2020neural}}}.
In closely related work, Zhou et al.~\shortcite{zhou2018stereo} represent a static scene with multiplane images (MPIs) generated from stereo pairs. Once generated, MPIs can be reused to efficiently render a range of output views. {DeepView}~\cite{flynn2019deepview} generates MPIs from multiview images using learned gradient descent. While MPIs allow for efficient rendering of dynamically varying viewpoints, generating an MPI from input images can be computationally expensive. Furthermore, MPIs can only render a limited range of views, i.e., perceptible artifacts often appear when the output view is relatively far from the reference views. Mildenhall et al.~\shortcite{mildenhall2019local} mitigate this problem by building multiple MPIs from a broad set of input views, which are blended together to create the output view, albeit with significant computational overhead (due to multiple MPI generation/rendering steps and costly 3D convolutional networks).
{\revise{Martin-Brualla et al.~\shortcite{martin2018lookingood} does real-time neural view synthesis, but it requires textured 3D reconstruction from a performance capture system as input and is limited to human scenes. Aliev et al.~\shortcite{aliev2020neural} introduces neural point-based representation but it requires a raw point cloud as input and presents limited quality.}}

A prominent line of recent research relates to neural radiance fields (NeRFs)~\cite{mildenhall2020nerf}. NeRFs represent a scene as an implicit function encoded by multilayer perceptrons (MLPs). NeRF-inspired approaches have led to rapid advances in view synthesis; however, none of these recent works directly address the unique challenges of VR passthrough. Specifically, many NeRF-derived methods rely on a dense set of views. More significantly, these methods incur prohibitive computational costs for per-scene optimization and volume rendering. {\revise{Despite recent work to decrease these costs~\cite{liu2020neural, yu2021plenoctrees, reiser2021kilonerf,hedman2021baking, garbin2021fastnerf}, NeRF-based methods still do not appear practical for VR passthrough. Similarly, other works have attempted to reduce the number and range of input views~\cite{yu2021pixelnerf, zhang2020nerf++, jain2021putting}, to support dynamic scenes~\cite{pumarola2021d, gao2021dynamic} and to be generalizable~\cite{yu2021pixelnerf,guo2022fast}; however, these lines of research have also fallen short on the crucial real-time requirement.}}

%% file: method.tex
\section{Neural Passthrough}
\label{sec:method}

In this section, we describe our neural passthrough system, including both optimizing the configuration of our hardware prototype (\secref{sec:method:hardware}) and the aspects of designing and training our real-time neural view synthesis method (Sections~\ref{sec:method:algorithm} and \ref{sec:method:training}, respectively).

\subsection{Optimizing the Hardware Configuration}
\label{sec:method:hardware}

As outlined in~\secref{sec:intro}, we focus on constructing a passthrough capture system with minimal hardware (i.e., using a pair of stereo RGB cameras as the only input). As such, the hardware design challenge can be posed as an optimization problem: where should the cameras be placed to best support view synthesis algorithms?

We define the objective as {\em{maximizing the  information captured from the 3D scene, by the stereo cameras, that is necessary for reconstructing the target novel views}}. In other words, we want to select the stereo camera placement to minimize the extent of the {\em{disocclusion}} regions (i.e., the set of 3D points that would be visible in the target novel views but that are occluded in the input views and, thus, cannot be faithfully recovered by view synthesis).

\begin{figure}[t]
\centering
\includegraphics[width=0.6\columnwidth]{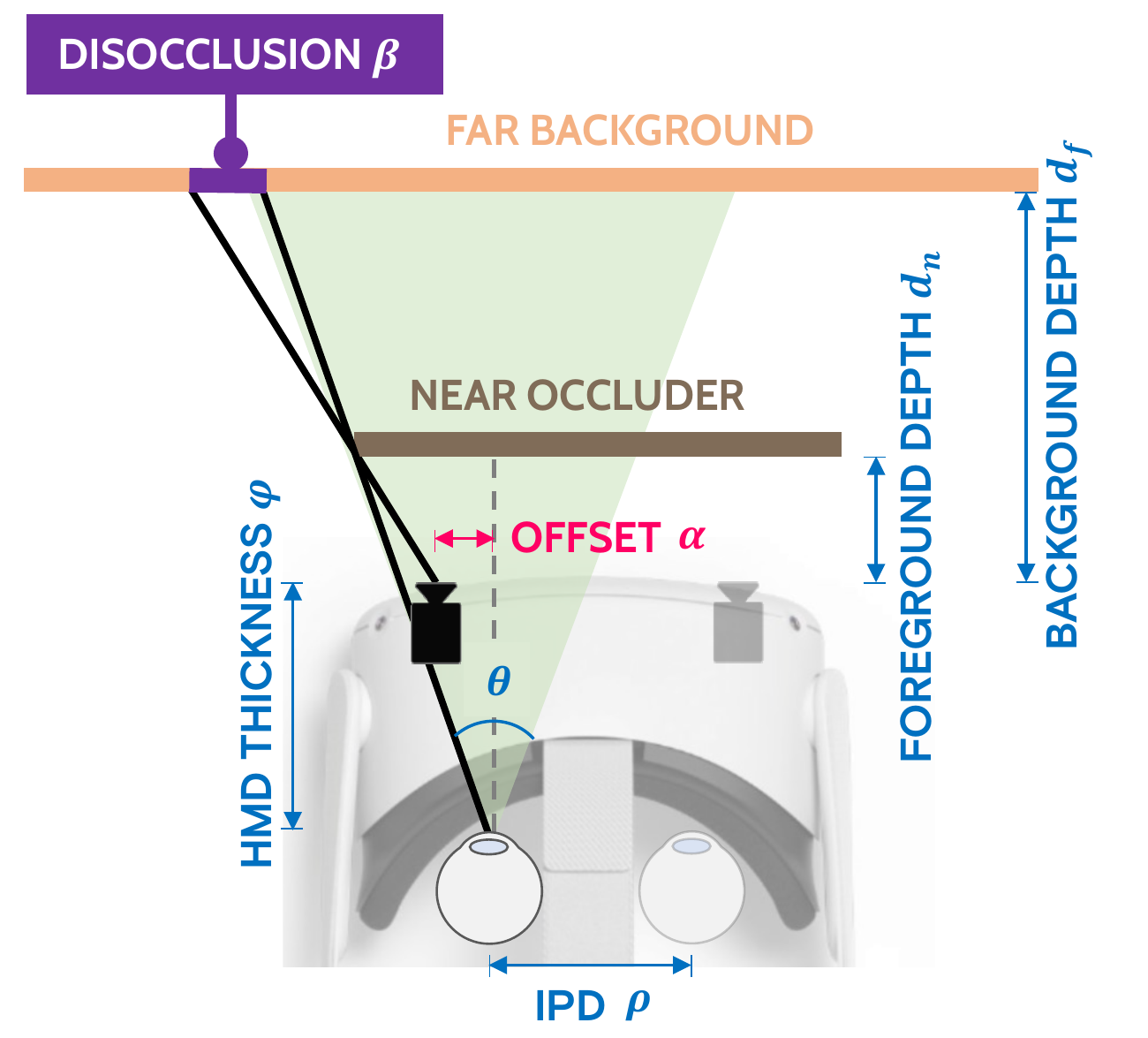}
\caption{The extent of \emph{disocclusion} regions (i.e., scene points that should be seen by the user's eyes but are occluded from the camera's perspective) depends on the VR headset construction, the position of the user's eyes, and the physical scene geometry. In this simplified model, the width $\beta$ of the disocclusion region is determined for a scene containing a near-field occluder and a far background plane. See Section~\ref{sec:method:hardware} for an application of this model to the optimization of stereo camera baselines for VR passthrough.}
\label{fig:disocc_analysis_diagram}
\end{figure}

In~\figref{fig:disocc_analysis_diagram}, we analyze the parameters that affect disocclusion. Note that, since we only consider a pair of cameras on the front surface of a HMD, we constrain both cameras to be vertically located on the same plane as the nominal eye positions and the optical centers of the viewing optics. Both cameras face directly forward and are further constrained to be symmetric about the HMD center. Under these constraints, the free parameters defining the camera placement reduce to just the horizontal offset $\xoffset$ between each camera and its corresponding eye. Intuitively, one might consider setting $\xoffset$ to zero, minimizing the distance between the input and target viewpoints. However, we propose to \emph{increase} $\xoffset$ (to a certain extent) to reduce the degree of disocclusions and, thus, provide some assistance to the view synthesis algorithm.

Applying the model in~\figref{fig:disocc_analysis_diagram}, disocclusion appears in the target view due to the viewpoint difference between the camera and the eye. The width of the disocclusion region $\disocclusionsize$ is given by
\noindent
\begin{eqnarray}
\begin{aligned}
\label{eq:disoc_size}
\disocclusionsize = \max\left(0, \hmdthickness \tan\left({\frac{\fieldofinterest}{2}}\right) - \xoffset\right) \cdot \left(\frac{\fardepth}{\neardepth} - 1\right)
\end{aligned}
\end{eqnarray}
\noindent where $\hmdthickness$ denotes the distance between the camera and the eye along the depth axis (i.e., the HMD thickness), $\neardepth$ and $\fardepth$ denote the depth of the near occluder and the background respectively ($\neardepth<\fardepth$), and $\fieldofinterest\in[0, \pi)$ measures the angular region within which the disocclusion is aimed to be eliminated. Note that, under our stereo camera constraints, only horizontal disocclusion can be reduced.
From Eq.~\eqref{eq:disoc_size}, clearly, when $\xoffset\geq\hmdthickness \tan{\frac{\fieldofinterest}{2}}$, disocclusion $\disocclusionsize$ will disappear. Given $\ipd$ is the target interpupillary distance (IPD), the required minimal stereo camera baseline becomes $\ipd + 2\,\hmdthickness \tan{\frac{\fieldofinterest}{2}}$.

From Eq.~\eqref{eq:disoc_size}, we also note that
reducing HMD thickness $\hmdthickness$ could reduce disocclusion $\disocclusionsize$. This suggests that the passthrough problem can benefit from more compact headset designs, such as those introduced by Maimone and Wang~\shortcite{maimone2020holocake}. In addition, disocclusion $\disocclusionsize$ increases when foreground objects are closer.

While this simplified model motivates slightly increasing the camera baseline in comparison to the user's IPD, this design choice does potentially introduce other issues. Following~\figref{fig:disocc_analysis_diagram}, as the left-hand camera moves to the left, it may reduce or even eliminate disocclusion on the left side of the occluder. However, disocclusion may be introduced on the right side. We observe that, since we have a stereo camera pair, increasing disocclusion in this manner may be somewhat compensated by utilizing observations from the other camera. Many anticipated near-field objects in VR use cases, such as hands and handheld implements, are compact enough such that the other camera can contribute to observing what would otherwise become a hidden surface. Thus, we have elected to increase our camera baseline in our prototype HMD in this manner. {\revise{Experimental results are given in the supplementary material.}}

As shown in~\figref{fig:teaser}, our prototype includes stereo RGB cameras (from Azure Kinect DK~\shortcite{AzureKinectDK}) attached to an Oculus Rift S headset~\shortcite{RiftS}.
The camera enclosures are removed and some electrical components are folded away from the front of the headset. We place the stereo cameras on a linear translation stage, allowing the baseline to be adjusted. The supported baseline ranges from 5.4cm to 10cm.  For the results shown in this paper, we selected a 10cm baseline. That supports an angular region $\fieldofinterest=25^\circ$, where the disocclusion is eliminated for an IPD of $\ipd=6\textrm{cm}$, or, equivalently, $\fieldofinterest=18^\circ$ for an IPD of $\ipd=7\textrm{cm}$. The distance between the cameras and the eyes (along the depth axis) is $\hmdthickness=9.3\textrm{cm}$. Each RGB camera runs at 30Hz with a resolution of $1280$$\times$$720$ and a $90^\circ$ field of view.

\begin{figure}[t]
\centering
\includegraphics[width=\columnwidth]{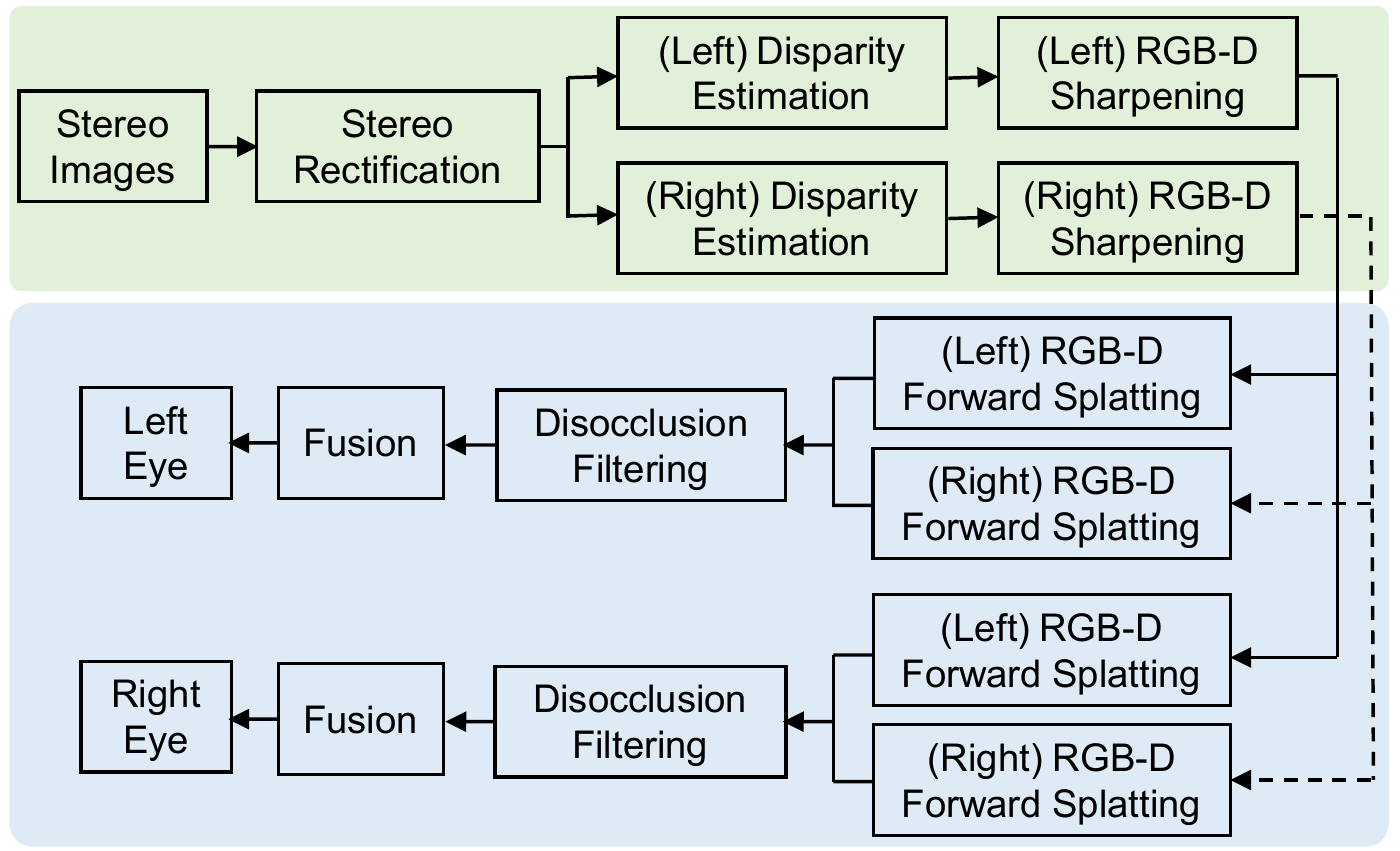}
\caption{Overview of the \emph{NeuralPassthrough} algorithm.}
\label{fig:method_overview}
\end{figure}

\subsection{Learned, Real-Time View Synthesis}
\label{sec:method:algorithm}

In this section, we define the {\em{NeuralPassthrough}} algorithm. This method addresses passthrough as an image-based rendering problem, solved separately per frame, taking stereo camera images as input and producing stereo images for the target eye views.

A block diagram of the method is provided in~\figref{fig:method_overview}. At a high level, the method represents the scene with 2D color and depth (RGB-D) images. A depth map is estimated at {\em{each}} of the input views by deep-learning-based disparity estimation (\secref{sec:method:algorithm:depth_estimation}). The RGB-D pixels of both input views are then splatted to each target view (\secref{sec:method:algorithm:forwardsplat}) before being fed into a neural network for final view reconstruction (\secref{sec:method:algorithm:fusion}). To reduce splatting artifacts due to the ambiguity of depth at discontinuities (e.g., ``flying'' pixels), the method filters the RGB-D data at each input view (\secref{sec:method:algorithm:rgbdsharpen}) before the splatting operation. The method further applies processing to reduce disocclusion artifacts in the splatted RGB values (\secref{sec:method:algorithm:disocclusion_filtering}) before passing them to the final reconstruction.

\begin{figure}
\centering
\includegraphics[width=\columnwidth]{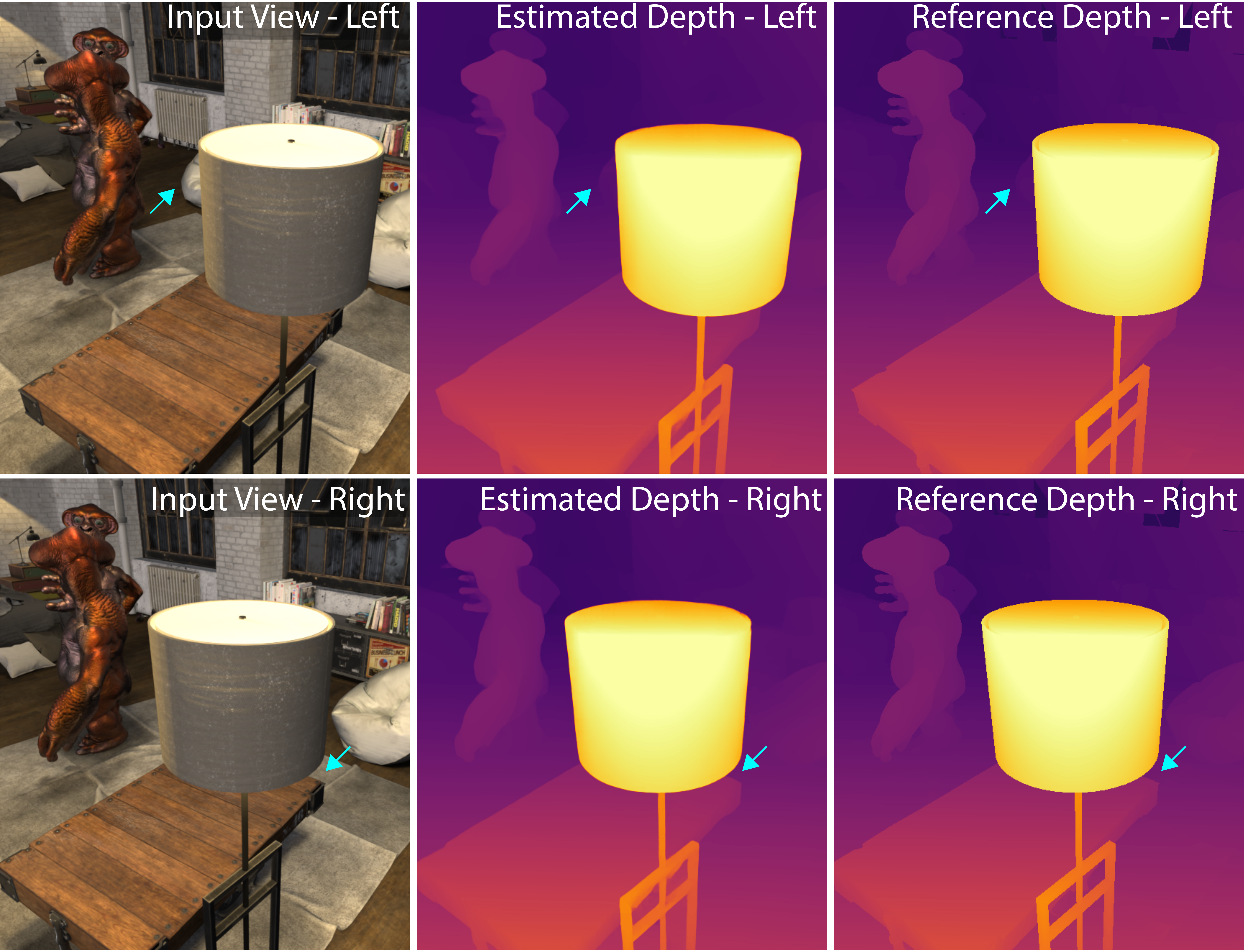}
\caption{Depth estimation example. Note that the depths, estimated by the algorithm in Section~\ref{sec:method:algorithm:depth_estimation}, well approximate the reference depths. The arrows indicate regions that are only visible in one of the stereo inputs, but where plausible depth is estimated from monocular depth cues.}
\label{fig:depth_est_example}
\end{figure}

\begin{figure}
\centering
\includegraphics[width=\columnwidth]{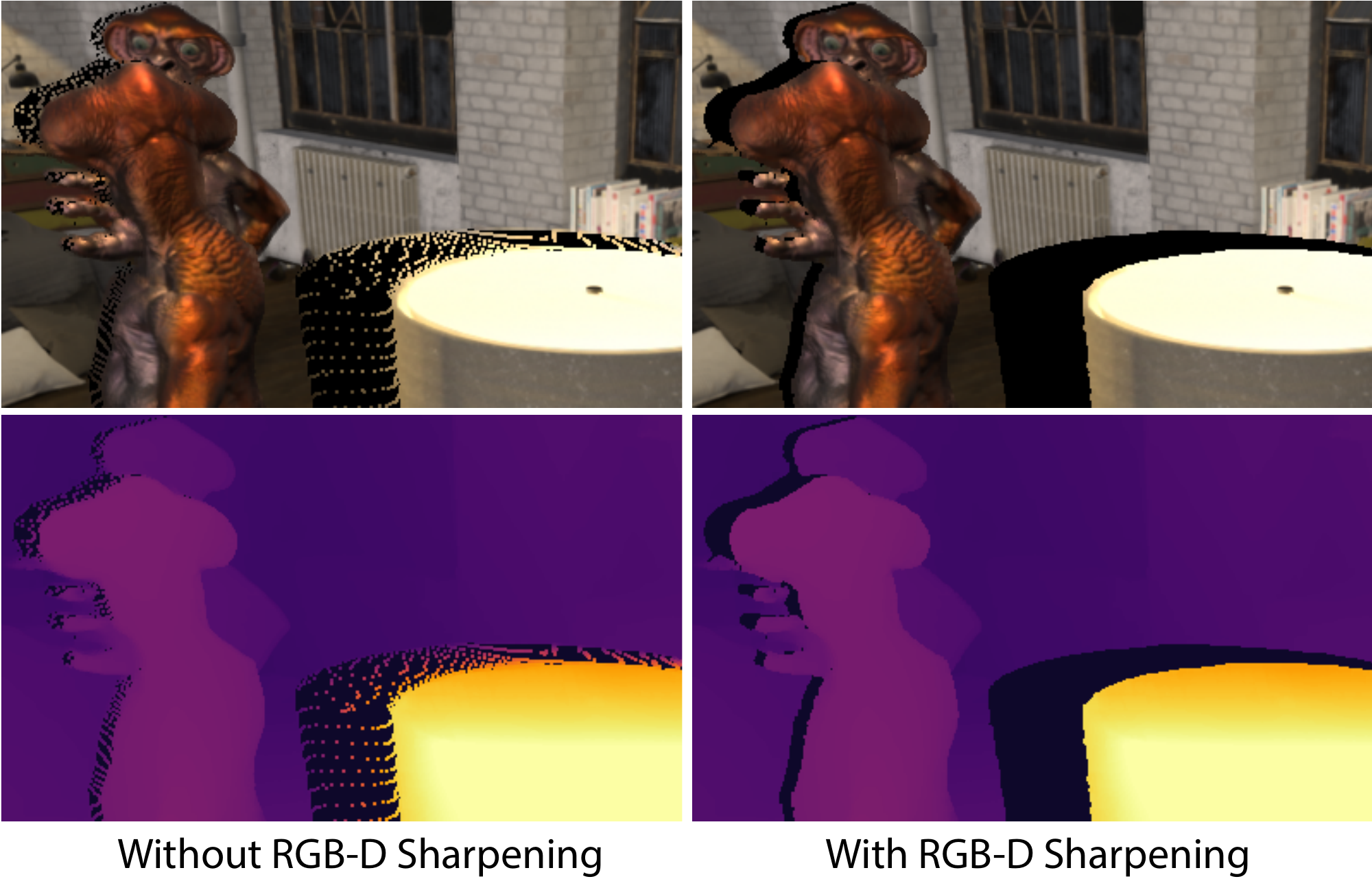}
\caption{Applying the RGB-D sharpening operation, following \secref{sec:method:algorithm:rgbdsharpen}, suppresses ``flying-pixel'' artifacts and produces cleaner depth maps. As shown on the left-hand side, without RGB-D sharpening, ``flying pixels'' are produced within the disocclusion regions (i.e., the background areas surrounding the edge of the lamp shade and the boundaries of the characters).}
\label{fig:flyingpixels}
\end{figure}

\subsubsection{Depth Estimation}
\label{sec:method:algorithm:depth_estimation}

We first rectify the input color image pairs, reducing disparity estimation from a 2D correspondence-matching problem to a more efficient 1D matching problem.
In contrast to \emph{Passthrough+}~\cite{chaurasia2020passthrough+}, which estimates scene depth from motion vectors produced by video encoding hardware, we leverage neural approaches to produce higher quality depth maps. Specifically, we apply the \emph{RAFT-Stereo} algorithm~\cite{lipson2021raft} to estimate a disparity map at each of stereo input views, which are then converted to inverse depth maps using pre-calibrated camera parameters. {\revise{We denote the rectified input color and the estimated inverse depth as $ \{\image_l, \image_r\}$ and $\{\sdepth_l, \sdepth_r\}$ respectively, where the subscript $l$ and $r$ indicate left and right views respectively.}}

~\figref{fig:depth_est_example} shows an example of the estimated depth maps recovered from stereo inputs, which accurately approximate the reference depth maps. Importantly, for regions that are only visible in one of the input views, the depth estimation network can still produce reasonable results from neighboring pixels and from monocular depth cues learned during training — differing from the motion-vector-based depth in \emph{Passthrough+}~\cite{chaurasia2020passthrough+} and the plane-sweep-volume approach in MPI-based methods~\cite{zhou2018stereo}. This is one of the key reasons that we choose to estimate the depth at {\em{each}} input view, since the two depth maps provide complementary information of the scene geometry.

\subsubsection{RGB-D Sharpening}
\label{sec:method:algorithm:rgbdsharpen}

While the estimated depth maps visually align with the corresponding color images, if they were directly used for view reprojection (\secref{sec:method:algorithm:forwardsplat}), ``flying pixels'' would occur at the disoccluded regions in the reprojected images, due to depth ambiguity at depth discontinuities (see~\figref{fig:flyingpixels}). To reduce this problem, we propose sharpening the color images and estimated depth maps along depth discontinuities. {\revise{Specifically, we detect the depth edges with Sobel filter followed by morphological dilation, and then set the RGB-D values of the edge pixels to that of their nearest-neighbor, non-edge pixels.}} We emphasize that another benefit of such RGB-D sharpening is that it helps produce clean depths in the splatted image space, which are important for the following disocclusion filtering step to work properly (\secref{sec:method:algorithm:disocclusion_filtering}).

\subsubsection{Forward Splatting}
\label{sec:method:algorithm:forwardsplat}

We elect to apply a neural network to reconstruct the color image, at each target eye viewpoint, from the color and recovered depth for the input stereo views. To reduce the required receptive field of the neural network, we first warp each input view to the target view. Since the depths are estimated for the input views, forward warping is required. Compared to backward warping, forward warping is more prone to introducing holes due to disocclusion; similarly, with forward warping, multiple source pixels can map to the same pixel in the warped image space, due to newly introduced occlusions. Both failure cases often occur for VR passthrough applications. In this section, we focus on the issue caused by newly introduced occlusions, leaving the discussion regarding disocclusion holes until~\secref{sec:method:algorithm:disocclusion_filtering}.

Fortunately, we have the estimated depth at each input view, providing visibility information for each 3D point. We employ the \emph{softmax} splatting method that was originally developed for video frame interpolation~\cite{niklaus2020softmax}. This method blends the pixels that were mapped to the same target pixel, applying pixel-wise importance weights defined as a measure of occlusion. In our implementation, we define the importance weights $\softsplatweight$ to be a function of the estimated inverse depth $\sdepth$, given by
\noindent
\begin{eqnarray}
\begin{aligned}
\label{eq:softsplatweight}
\softsplatweight = 36 \left(\frac{\sdepth - \sdepthmin}{\sdepthmax - \sdepthmin} + \frac{1}{9} \right)
\end{aligned}
\end{eqnarray}
\noindent where $\sdepthmin$ and $\sdepthmax$ are the minimum and maximum of the inverse depth map $\sdepth$, and the heuristic constants are chosen to map the weights to the range [4, 40], which works well in our experiments. The metric $\softsplatweight$ assigns higher weights to the source pixels closer to the cameras (in the warped image space). We denote the splatted color and inverse depth as $\{\warpedcolor_l,\warpedcolor_r\}$ and $\{\warpeddepth_l,\warpeddepth_r\}$, respectively.

\subsubsection{Disocclusion Filtering}
\label{sec:method:algorithm:disocclusion_filtering}

\begin{figure}[h]
\centering
\includegraphics[width=\columnwidth]{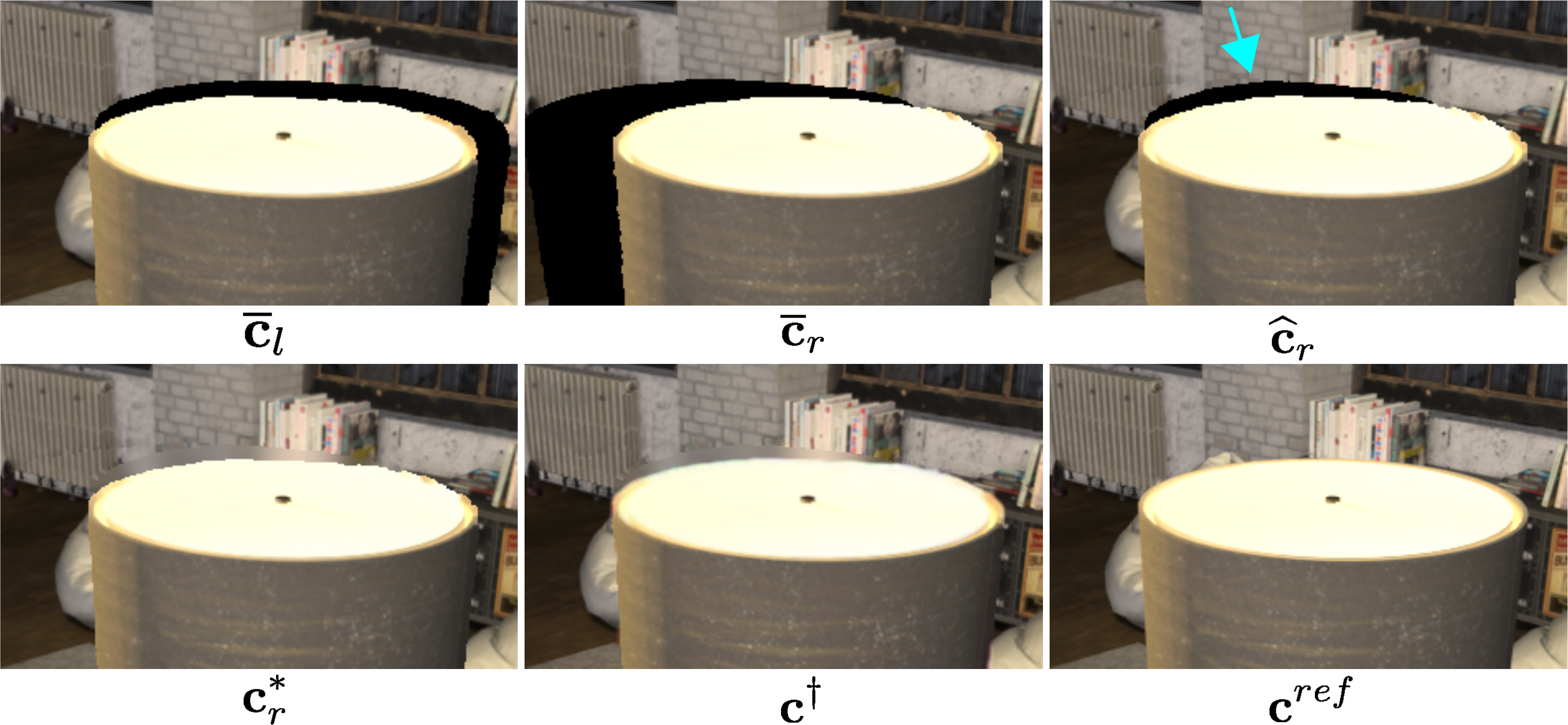}
\caption{Example results for disocclusion filtering, as defined in~\secref{sec:method:algorithm:disocclusion_filtering}. $\warpedcolor_l$ and $\warpedcolor_r$ are the forward-splatted color images, as taken from the input left and right view, respectively (see~\secref{sec:method:algorithm:forwardsplat}). $\partoccfilteredcolor_r$ and $\fulloccfilteredcolor_r$ are the right view after partial (Eq.~\eqref{eq:partialocclusionfill}) and full disocclusion filtering (Eq.~\eqref{eq:fullocclusionfill}), respectively. $\finalcolor$ is the final reconstruction (Eq.~\ref{eq:fusion}), and $\gtcolor$ is the reference for the target view. The arrows point to regions with full disocclusion holes.}
\label{fig:disocclusion_filtering}
\end{figure}

The forward-splatted images at the target views typically contain holes due to disocclusions, as shown in~\figref{fig:disocclusion_filtering}. In this section, we describe our approach to address this issue. We divide the disocclusion holes into two categories and treat them separately: {\em{partial disocclusion}}, defined as the holes that occur in only one of the splatted images (i.e., either $\warpedcolor_l$ or $\warpedcolor_r)$, and {\em{full disocclusion}}, defined as the holes that occur in both $\warpedcolor_l$ and $\warpedcolor_r$.

Partial disocclusion can be removed by blending $\warpedcolor_l$ and $\warpedcolor_r$:
\noindent
\begin{eqnarray}
\begin{aligned}
\label{eq:partialocclusionfill}
\partoccfilteredcolor_l &= (\V{1} - \occlusionmask_l) \odot \warpedcolor_l + \occlusionmask_l \odot \warpedcolor_r \\
\partoccfilteredcolor_r &= (\V{1} - \occlusionmask_r) \odot \warpedcolor_r + \occlusionmask_r \odot \warpedcolor_l
\end{aligned}
\end{eqnarray}
\noindent where $\odot$ denotes the Hadamard product, and the pixel-wise masks $\occlusionmask_l$ and $\occlusionmask_r$ are defined on the splatted inverse depth $\warpeddepth_l$ and $\warpeddepth_r$, as 
\noindent
\begin{equation}
\label{eq:occlusionmask}
\occlusionmask_l =
    \begin{cases}
      1 & \text{if}\ \warpeddepth_l < \epsilon\\
      0 & \text{otherwise}
    \end{cases}
,\quad
\occlusionmask_r =
    \begin{cases}
      1 & \text{if}\ \warpeddepth_r < \epsilon\\
      0 & \text{otherwise}
    \end{cases}
\end{equation}
\noindent where $\epsilon=0.1$ and $\{\occlusionmask_l, \occlusionmask_r\}$ indicate the zero-valued pixels in the splatted inverse depth maps $\{\warpeddepth_l, \warpeddepth_r\}$. Partial disocclusion removal results are shown in~\figref{fig:disocclusion_filtering}.

Full disocclusions can not be faithfully recovered as the input stereo inputs contain no information for those regions. As reviewed in Section~\ref{sec:related:learning}, prior work in 3D photography~\cite{shih3d} resolves disocclusions via context-aware image inpainting. We find that for {\em{static}} images visually acceptable results can be hallucinated; but, if applied to our \emph{dynamic} passthrough problem, this approach will introduce temporal flickering into the output videos. Furthermore, advanced image inpainting techniques~\cite{shih3d,liu2018image} are typically computationally expensive, hindering their utility for our real-time application.

\begin{algorithm}[t]
	\caption{Full Disocclusion Filtering}
	\label{alg:occlusion_filter}
	\SetAlgoNoLine
\KwIn{Color image $\partoccfilteredcolor$, inverse depth $\warpeddepth$, occlusion mask $\fullocclusionmask$, kernel $\psf$}
\KwOut{Filtered color image $\fulloccfilteredcolor$}
\begin{algorithmic}
\FOR{{each pixel} i}
      \IF{$\fullocclusionmask(i)$ is 0}
        \STATE ${\fulloccfilteredcolor(i) = \partoccfilteredcolor(i)}$
      \ELSE
        \STATE {$\warpeddepth^{\mathcal{N}}_{{min}},\warpeddepth^{\mathcal{N}}_{{max}}, c_{acc}, w_{acc}$ = {MAX, MIN, 0, 0}}
        \FOR{each pixel j in local neighborhood $\mathcal{N}_{i}$}
           \IF{$\warpeddepth(j)>0.01$}
              \STATE $\warpeddepth^{\mathcal{N}}_{{min}}$, $\warpeddepth^{\mathcal{N}}_{{max}}$ = $min(\warpeddepth^{\mathcal{N}}_{{min}}, \warpeddepth(j))$, $max(\warpeddepth^{\mathcal{N}}_{{max}}, \warpeddepth(j))$
           \ENDIF
        \ENDFOR
        \FOR{each pixel j in local neighborhood $\mathcal{N}_{i}$}
            \IF{$\warpeddepth(j)>0.01$ \AND $\warpeddepth(j)<0.5 (\warpeddepth^{\mathcal{N}}_{{min}}$ + $\warpeddepth^{\mathcal{N}}_{{max}})$}
               \STATE $c_{acc} \pluseq \partoccfilteredcolor(j) \cdot \psf(i,j)$
               \STATE $w_{acc} \pluseq \psf(i,j)$
            \ENDIF
        \ENDFOR
        \IF{$w_{acc} > 0$}
           \STATE $\fulloccfilteredcolor(i) = {c_{acc}} / {w_{acc}}$
        \ELSE
           \STATE {$\fulloccfilteredcolor(i) = \partoccfilteredcolor(i)$}
        \ENDIF
      \ENDIF
\ENDFOR
\end{algorithmic}
\end{algorithm}

As an alternative, we propose depth-assisted, anisotropic low-pass filtering to produce stable results efficiently. We observe that the disoccluded regions are more often missing information from background objects rather than from foreground occluders; as a result, our method fills in the disoccluded pixels using only the smoothed colors of {\em{relatively far}} objects within the local neighborhood. The method details are given by in Eq.~\eqref{eq:fullocclusionfill} and Algorithm~\ref{alg:occlusion_filter}:
\begin{eqnarray}
\begin{aligned}
\label{eq:fullocclusionfill}
\widehat{\occlusionmask} &= \occlusionmask_l \odot \occlusionmask_r \\
\fulloccfilteredcolor_l &= {\mfulldisocclusionfiltering}(\partoccfilteredcolor_l, \warpeddepth_l, \fullocclusionmask, \psf) \\
\fulloccfilteredcolor_r &= {\mfulldisocclusionfiltering}(\partoccfilteredcolor_r, \warpeddepth_r, \fullocclusionmask, \psf)
\end{aligned}
\end{eqnarray}

\noindent The mask $\widehat{\occlusionmask}$ indicates whether a pixel is fully disoccluded. $\psf$ denotes a low-pass kernel (which in our implementation is a zero-mean 2D Gaussian filter with size $29\times29$ and a standard deviation of 7 pixels). For experimental evaluation of the benefits of the filtering operations, see Table~\ref{tab:result_synthetic}.

\subsubsection{Fusion}
\label{sec:method:algorithm:fusion}

Our pipeline concludes by feeding the disocclusion-filtered images to a neural network for final reconstruction at the target eye views,  as given by
\begin{eqnarray}
\begin{aligned}
\label{eq:fusion}
\finalcolor = {\mfusion}(\fulloccfilteredcolor_l, \fulloccfilteredcolor_r) \\ %
\end{aligned}
\end{eqnarray}
\noindent where the fusion network is a lightweight U-Net with skip connections, comprising the specific architecture in Table~\ref{tab:fusion_network}.
Note that the fusion network runs once for each of the two target eye views, as illustrated in~\figref{fig:method_overview}. We find that fusion is necessary to further reduce reprojection errors and aliasing artifacts in $\fulloccfilteredcolor_l$ and $\fulloccfilteredcolor_r$.

 \begin{table}
 \centering
 \caption{ Each layer of the fusion network (Section~\ref{sec:method:algorithm:fusion}) is a 2D convolution followed by relu activation. The operators concat, down, and up, represent concatenation, average pooling, and bilinear upsampling, respectively.}
 \label{tab:fusion_network}
     \begin{tabular}{p{1ex}p{10ex}llc}
 \hline
 & Layer & Input Tensor & Channels In/Out \\
 \hline
     & conv0        & concat($\fulloccfilteredcolor_l$, $\fulloccfilteredcolor_r$) & 6/16 \\
     & conv1        & conv0  & 16/16\\
     & conv2        & down(conv1) & 16/32 \\
     & conv3         &  conv2 & 32/32\\
     & conv4        & down(conv3) & 32/64 \\
     & conv5        &  conv4 & 64/64 \\
     & conv6        & concat$\bigl($up(conv5), conv$\bigr)$) & 96/32 \\
     & conv7        &  layer6 & 32/32\\
     & conv8        & concat$\bigl($up(conv7), conv1$\bigr)$  & 48/16  \\
     & conv9       &  conv8 & 16/16 \\
     & conv10       &  conv9 & 16/3 \\
 \hline
 \end{tabular}
 \end{table}

\begin{table}
	\centering
	\caption{Quality comparisons for \emph{NeuralPassthrough} (Ours) on synthetic datasets. Average PSNR (dB), SSIM and ST-RRED are reported. Higher PSNR and SSIM, and lower ST-RRED indicate better quality.}
	\label{tab:result_synthetic}
	\begin{tabular}{lllll}
		\hline
		&  & PSNR$\uparrow$ & SSIM$\uparrow$ & ST-RRED$\downarrow$  \\
		\hline
		&MPI ~\cite{zhou2018stereo} & 27.38  & 0.8818 & 105.74 \\
		&Ours     & {\bf{30.74}}  & {\bf{0.9579}} & {\bf{51.78}} \\
		&Ours (trained without Eq.~\ref{eq:fullocclusionfill})     & 28.66  & 0.9475 & 95.33 \\
		&Ours (trained without Eq.~\ref{eq:partialocclusionfill},~\ref{eq:fullocclusionfill})     & 29.02  & 0.9456 & 99.33 \\
		\hline
	\end{tabular}
\end{table}

\subsection{Training}
\label{sec:method:training}

The training loss function for \emph{NeuralPassthrough} is defined as
\noindent
\begin{eqnarray}
\begin{aligned}
\label{eq:loss}
10\,|| (1 - \fullocclusionmask) \odot (\finalcolor - \gtcolor)||_1 - ||(1 - \fullocclusionmask) \odot \metricssim (\finalcolor, \gtcolor)||_1
\end{aligned}
\end{eqnarray}
\noindent where $\metricssim$ is the structural similarity index measure~\cite{wang2004image}. We apply the mask $(1 - \fullocclusionmask)$ to exclude the full disocclusion regions from the loss, preventing inpainting at those regions (which may lead to inconsistent temporal and left/right completions that, in turn, could degrade the user experience, especially due to the limited capacity (by design) of the fusion network in~\secref{sec:method:algorithm:fusion}). The stereo depth network reapplies the pre-trained \emph{RAFT-Stereo} model with frozen weights during training.

{\revise{We train the method on a synthetic dataset with 80 random scenes, similar to the ones in Xiao et al.~\shortcite{xiao2018deepfocus}. Each scene includes 3D scans of sculptures from the Louvre as well as spheres and cubes, that are randomly textured and placed in 3D space with varying depths. The unnatural geometry and appearance makes the training data substantially different from the synthetic and real scenes we test in the paper. Example scenes are provided in the supplementary material.}} For each scene, we render 20 image sequences with a resolution of $512\times512$ and rendered at varying viewpoints (i.e., two views serve as the input stereo pair, with a 10cm baseline, and the rest are the target output views that are 9.3cm behind the input views and with baselines ranging from 4.8cm to 8.0cm). Note that the trained network can be applied to other camera and IPD configurations, as well as to different input resolutions, at test time.
We train the method in Pytorch using the ADAM optimizer with default parameters for 120 epoches.

%% file: results.tex
\section{Results}
\label{sec:results}

\subsection{Real-Time Implementation}

After training, we implement our full, optimized inference method in C++ using CUDA/CuDNN, integrating it with the Oculus Rift SDK for real-time passthrough demonstrations. We test the method on a desktop with an Intel Xeon W-2155 CPU and dual Nvidia Titan V GPUs. Each GPU is responsible for one depth estimation and one eye-view reconstruction. The method runs at 32ms per frame at a resolution of $1280$$\times$$720$ for the two eye views, with depth estimation at 24.0ms, RGB-D sharpening at 0.3ms, forward-splatting at 0.8ms, disocclusion filtering at 0.9ms, and the fusion network at 5.6ms.

 \begin{figure}
\centering
\includegraphics[width=\columnwidth]{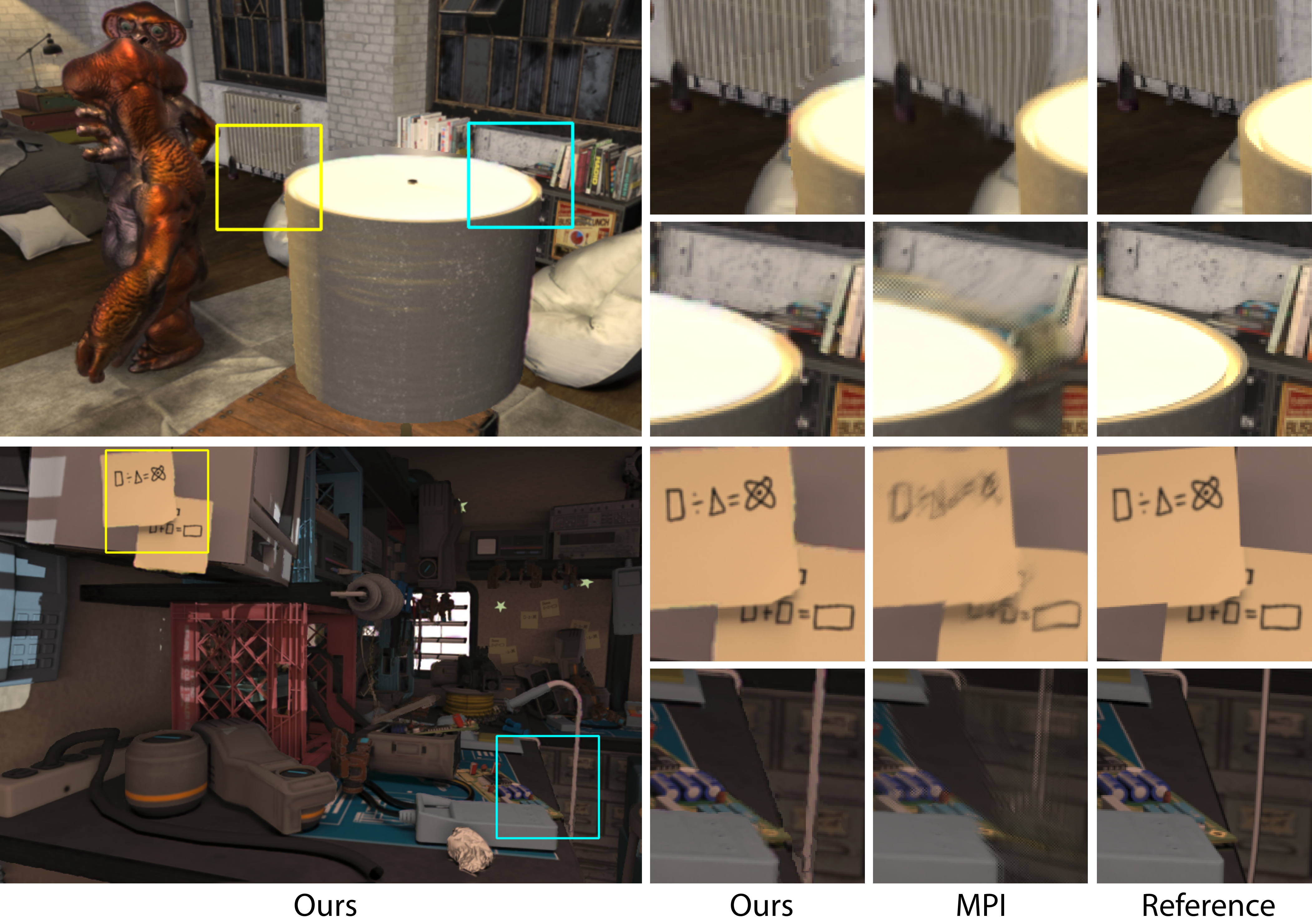}
\caption{Comparisons on synthetic data. For the \emph{DanceStudio} scene (top), our PSNR and SSIM are 34.28dB and 0.97, while MPI~\cite{zhou2018stereo} achieves 29.69dB and 0.90. For the \emph{ElectronicRoom} scene (bottom), ours are 30.06dB and 0.95, while MPI achieves 24.42dB and 0.86. The MPI results present obvious artifacts (e.g., stretching and repeated textures at disocclusions).}
\label{fig:result_synthetic}
\end{figure}

\begin{figure}
\centering
\includegraphics[width=\columnwidth]{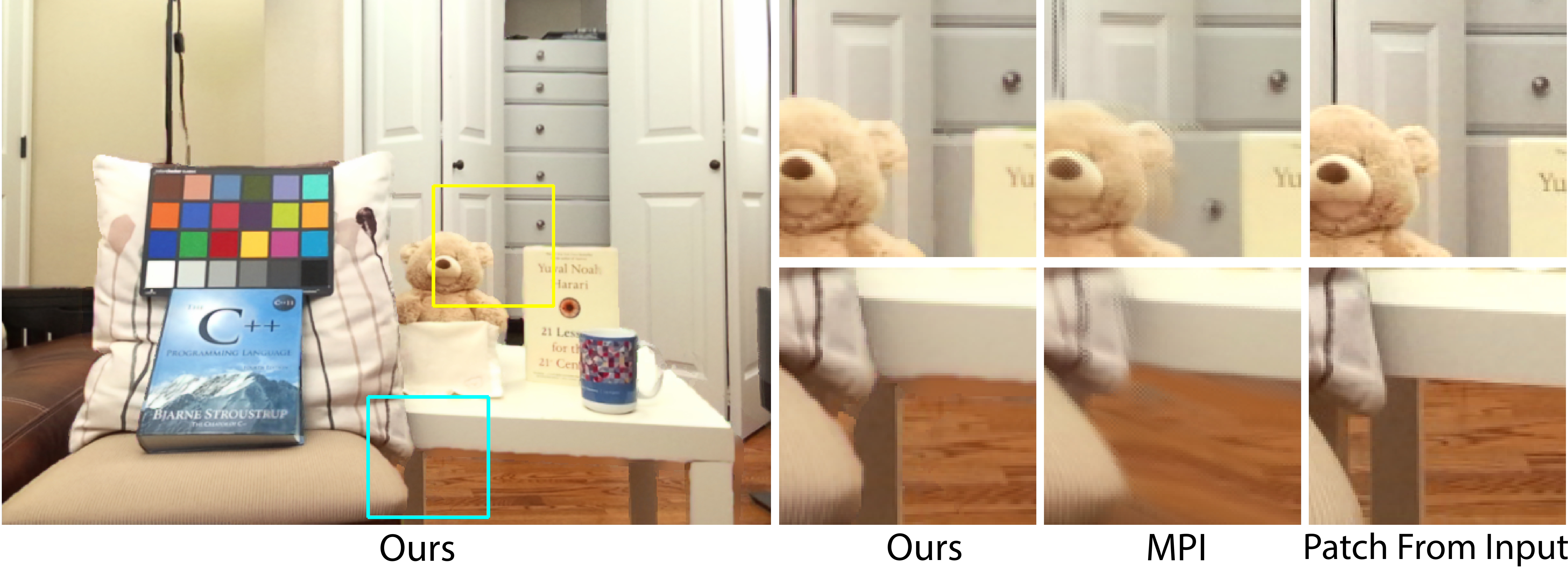}
\caption{Qualitative comparisons on data captured by our prototype headset. Our method does not produce texture stretching artifacts, while MPI~\cite{zhou2018stereo} tends to, while also failing to reconstruct the table leg.}
\label{fig:result_protoCC_static}
\end{figure}

While we report these run times for our per-frame processing pipeline, we see opportunities for further improvement. For example, we could run the depth estimation at a lower frame rate (e.g., 30Hz), but run the color reconstruction operations (i.e., forward splatting, disocclusion filtering, and fusion, which take 7.3ms in total) at a higher frame rate (e.g., $72$Hz or the native refresh rate of the displays). In this manner, the more efficient color reconstruction pipeline could account for the current estimate of the head pose, while reusing the latest depth estimates. Notably, such a \emph{decoupled} framework is similar to \emph{Passthrough+}, which reconstructs the textured meshes at $30$Hz, but renders the eye buffers at $72$Hz.

\begin{figure}
\centering
\includegraphics[width=\columnwidth]{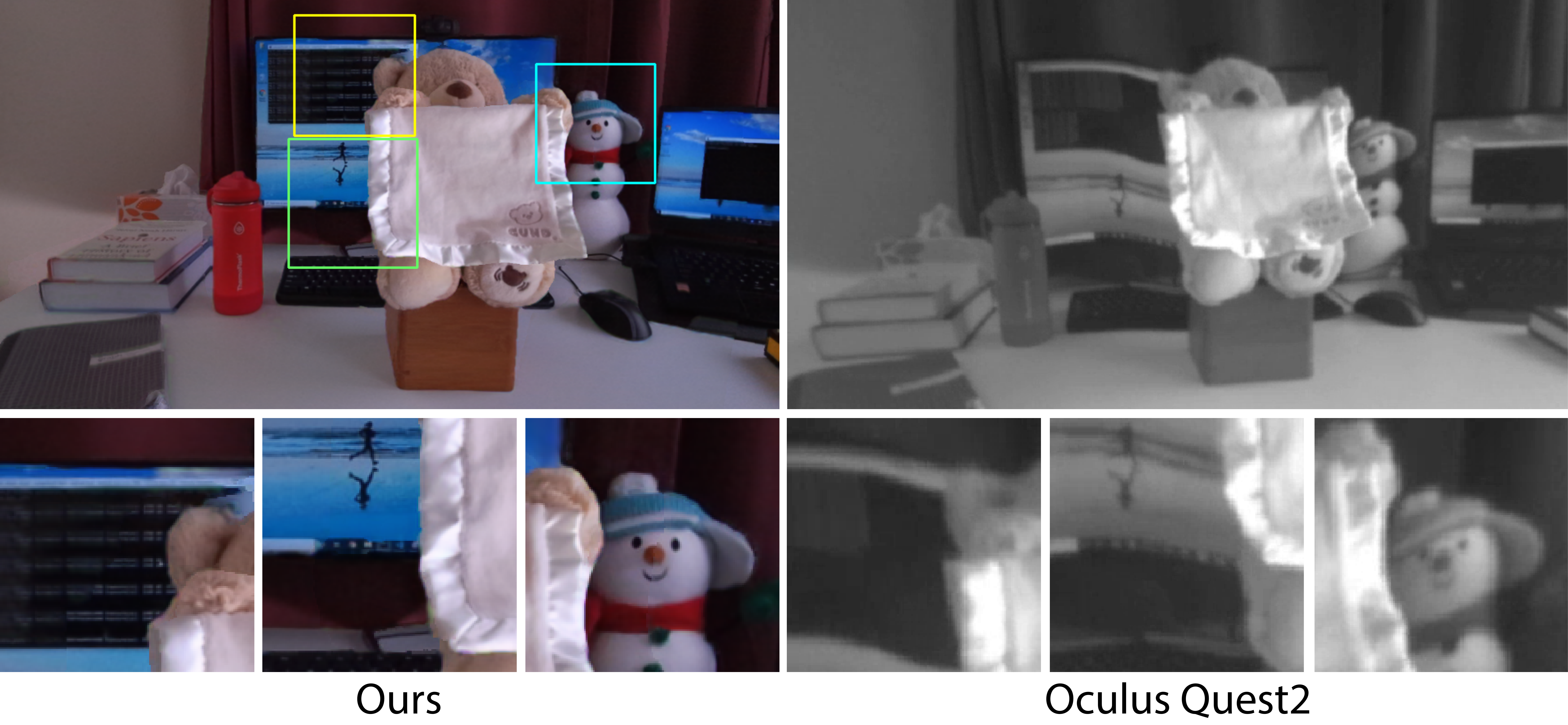}
\caption{Visual comparisons between our method and Oculus Quest 2, which is closely related to Passthrough+~\cite{chaurasia2020passthrough+}.}
\label{fig:result_quest2_comparison_desktopvideo}
\end{figure}

\begin{figure}
\centering
\includegraphics[width=\columnwidth]{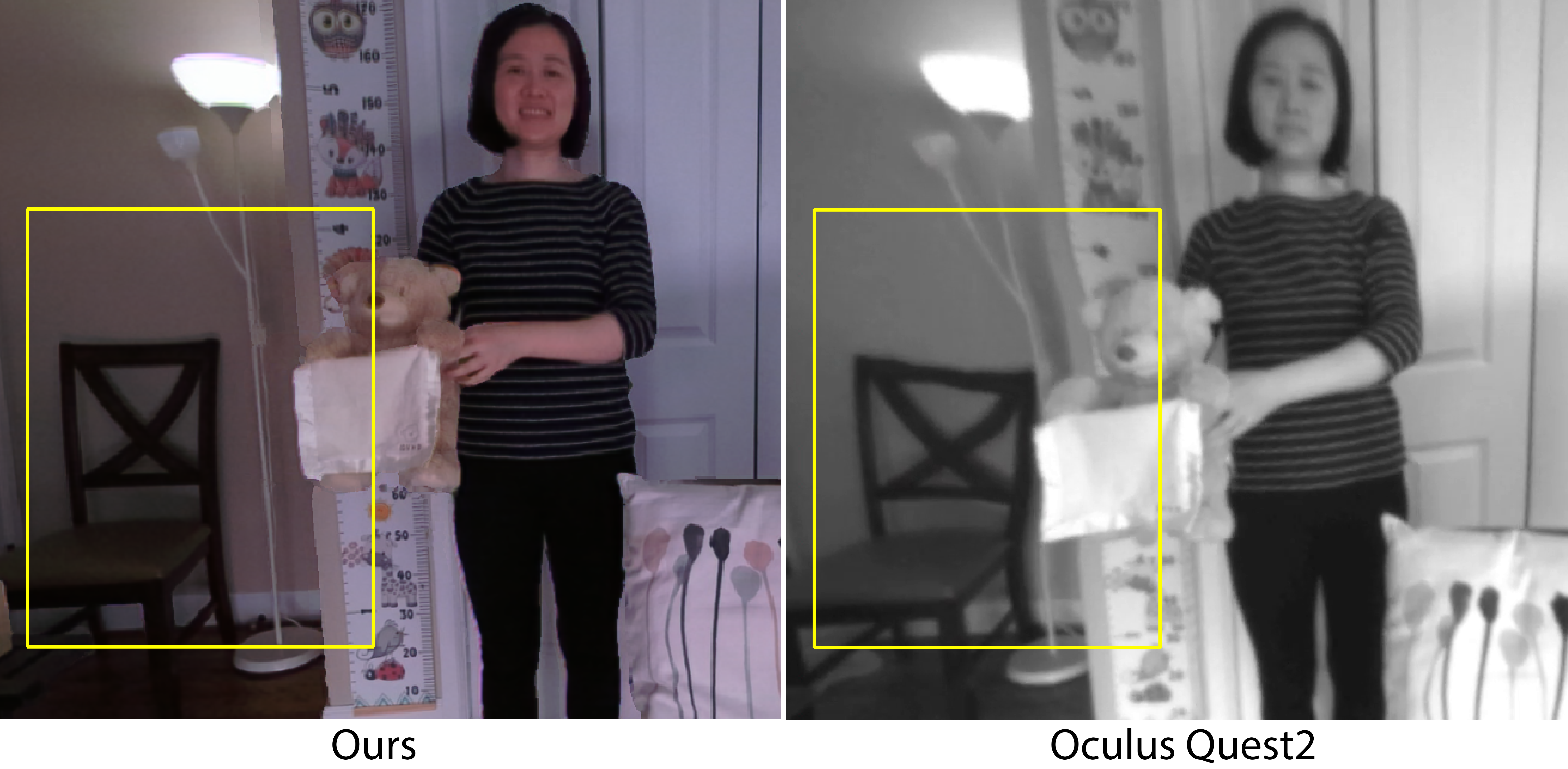}
\caption{Visual comparisons between our method and Oculus Quest 2 on a dynamic scene.
Our method does not show severe distortions, whereas Quest 2 does as highlighted.}
\label{fig:result_quest2_comparison_personvideo}
\end{figure}

\subsection{Quality Comparisons}
\label{sec:results:quality_comparisons}

As there is little recent work on real-time view synthesis, we compare to the representative MPI method~\cite{zhou2018stereo}, which also takes stereo images as inputs. MPI requires several seconds to generate the multiplane representation and another several seconds to render stereo eye views (at a resolution of $1280$$\times$$720$ using TensorFlow with our system). Although follow-up MPI works exist with improved quality~\cite{srinivasan2019pushing,mildenhall2019local}, these approaches are substantially slower due to the need to generate/render multiple MPIs per frame and the use of 3D convolutional networks, making them even less applicable to VR passthrough.

For image quality comparisons, we render two datasets for 3D environments with dynamic objects ({\em{DanceStudio}} and {\em{ElectronicRoom}}). Each dataset contains 5 videos with simulated VR head movements, each containing 30 frames at a resolution of $1280$$\times$$720$. For each frame, we render input stereo views (with a 10cm baseline) and target eye views (with a 6cm IPD and depth-axis offset of 9.3cm). These scenes differ in appearance from our static training datasets.

We evaluate the methods using PSNR, SSIM~\cite{wang2004image} and Spatio-Temporal Entropic Difference (ST-RRED)~\cite{soundararajan2012video}, where the latter is for video quality and temporal stability assessment. As reported in Table~\ref{tab:result_synthetic}, our method outperforms MPI by a large margin on all metrics. Example result are shown in~\figref{fig:result_synthetic}. Notably, MPI presents more obvious artifacts, especially distortion, stretching, and repeated textures at disocclusion regions.

We qualitatively compare these methods using real data captured by our prototype, as shown in~\figref{fig:result_protoCC_static}. Since we can not capture the ground truth reference images at the target eye views, we provide the closest patches from the input views as a visual reference.

As reviewed in Sections~\ref{sec:intro}, the closest related work to our \emph{NeuralPassthrough} is \emph{Passthough+}~\cite{chaurasia2020passthrough+}, which is the predecessor to the current commercial solution applied in Oculus Quest 2. As this commercial implementation is unavailable, and its camera images are not accessible via the Oculus SDK, we capture scenes with similar camera trajectories using our prototype and the Quest 2, allowing for qualitative visual comparisons. Example results are shown in Figures~\ref{fig:result_quest2_comparison_desktopvideo} and~\ref{fig:result_quest2_comparison_personvideo}. We observe that the main limitation of Oculus Quest 2 is that the reconstructed mesh can be inaccurate at depth discontinuities and within disocclusion regions, causing noticeable distortion and stretching artifacts. In contrast, our method produces more accurate results, while also allowing for color outputs with higher resolution. Please refer to the supplementary material for video results and comparisons.

%% file: limitations.tex
\section{Limitations and Future Work}
\label{sec:limitations}

The quality of our results is partly affected by the quality of our real-time depth estimation. While the depth estimation module produces reasonable results in most circumstances, it may fail for objects with challenging geometric details, for view-dependent materials, or when the monocular depth cues are lacking. Our results may also present blending artifacts when the input stereo views contain mismatched colors due to severe view-dependent reflections, especially when they appear together with disocclusion.
Examples of these failure cases are shown in~\figref{fig:failure_cases}.

Future improvements in real-time depth estimation may benefit our approach, updating the associated step of our pipeline. In future work, active depth sensors may provide instant depth generation and improve depth estimation for regions where stereo recovery does not work well (e.g. regions with little texture, difficult partial disocclusions, or view-dependent effects). To enable this work, our hardware prototype already incorporates depth sensors.

Oculus Quest 2 passthrough results suffer from severe distortion artifacts, however they typically appear temporally stable, partly due to their low resolution, but also due to their multi-frame reconstruction method. As future work, we anticipate leveraging multiple frames to further improve spatial quality and temporal consistency.

%% file: conclusion.tex
\section{Conclusion}
\label{sec:conclusion}

This paper takes a first step towards bringing the latest developments in neural view synthesis to the specific domain of VR passthrough. With VR recently seeing wider adoption, we believe improved passthrough technologies are necessary to unlock a broad set of mixed reality applications — seamlessly blending virtual objects with the user's physical surroundings. Throughout this paper we have emphasized that, while neural view synthesis is an increasingly well studied topic, VR applications set a much higher bar on performance. To deliver compelling VR passthrough, the field will need to make significant strides both in image quality (i.e., suppressing notable warping and disocclusion artifacts), while meeting the strict real-time, stereoscopic, and wide-field-of-view requirements. Tacking on the further constraint of mobile processors for wearable computing devices means that there truly is a long road ahead.

\begin{figure}
\centering
\includegraphics[width=\columnwidth]{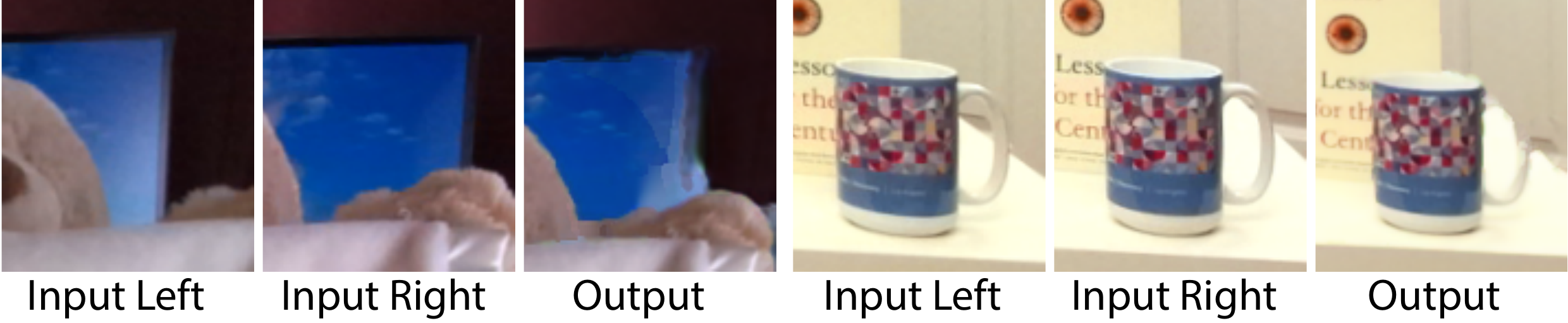}
\caption{Example failure cases on real data. Our system may produce artifacts for objects with significant view-dependent reflections (monitor corner) or with complex geometry and ambiguities in stereo matching (mug handle).}
\label{fig:failure_cases}
\end{figure}

%% file: main.bbl

\begin{thebibliography}{43}


\ifx \showCODEN    \undefined \def \showCODEN     #1{\unskip}     \fi
\ifx \showDOI      \undefined \def \showDOI       #1{#1}\fi
\ifx \showISBNx    \undefined \def \showISBNx     #1{\unskip}     \fi
\ifx \showISBNxiii \undefined \def \showISBNxiii  #1{\unskip}     \fi
\ifx \showISSN     \undefined \def \showISSN      #1{\unskip}     \fi
\ifx \showLCCN     \undefined \def \showLCCN      #1{\unskip}     \fi
\ifx \shownote     \undefined \def \shownote      #1{#1}          \fi
\ifx \showarticletitle \undefined \def \showarticletitle #1{#1}   \fi
\ifx \showURL      \undefined \def \showURL       {\relax}        \fi
\providecommand\bibfield[2]{#2}
\providecommand\bibinfo[2]{#2}
\providecommand\natexlab[1]{#1}
\providecommand\showeprint[2][]{arXiv:#2}

\bibitem[Azu(2021)]%
        {AzureKinectDK}
 \bibinfo{year}{2021}\natexlab{}.
\newblock \bibinfo{title}{Azure Kinect DK}.
\newblock
\newblock
\urldef\tempurl%
\url{https://azure.microsoft.com/en-us/services/kinect-dk/}
\showURL{%
\tempurl}


\bibitem[Rif(2021)]%
        {RiftS}
 \bibinfo{year}{2021}\natexlab{}.
\newblock \bibinfo{title}{Rift-S VR}.
\newblock
\newblock
\urldef\tempurl%
\url{https://www.oculus.com/rift-s/features/}
\showURL{%
\tempurl}


\bibitem[Aliev et~al\mbox{.}(2020)]%
        {aliev2020neural}
\bibfield{author}{\bibinfo{person}{Kara-Ali Aliev}, \bibinfo{person}{Artem
  Sevastopolsky}, \bibinfo{person}{Maria Kolos}, \bibinfo{person}{Dmitry
  Ulyanov}, {and} \bibinfo{person}{Victor Lempitsky}.}
  \bibinfo{year}{2020}\natexlab{}.
\newblock \showarticletitle{Neural point-based graphics}. In
  \bibinfo{booktitle}{\emph{European Conference on Computer Vision}}. Springer,
  \bibinfo{pages}{696--712}.
\newblock


\bibitem[Broxton et~al\mbox{.}(2020)]%
        {broxton2020immersive}
\bibfield{author}{\bibinfo{person}{Michael Broxton}, \bibinfo{person}{John
  Flynn}, \bibinfo{person}{Ryan Overbeck}, \bibinfo{person}{Daniel Erickson},
  \bibinfo{person}{Peter Hedman}, \bibinfo{person}{Matthew Duvall},
  \bibinfo{person}{Jason Dourgarian}, \bibinfo{person}{Jay Busch},
  \bibinfo{person}{Matt Whalen}, {and} \bibinfo{person}{Paul Debevec}.}
  \bibinfo{year}{2020}\natexlab{}.
\newblock \showarticletitle{Immersive light field video with a layered mesh
  representation}.
\newblock \bibinfo{journal}{\emph{ACM Transactions on Graphics (TOG)}}
  \bibinfo{volume}{39}, \bibinfo{number}{4} (\bibinfo{year}{2020}),
  \bibinfo{pages}{86--1}.
\newblock


\bibitem[Chapiro et~al\mbox{.}(2014)]%
        {chapiro2014optimizing}
\bibfield{author}{\bibinfo{person}{Alexandre Chapiro}, \bibinfo{person}{Simon
  Heinzle}, \bibinfo{person}{Tun{\c{c}}~Ozan Ayd{\i}n}, \bibinfo{person}{Steven
  Poulakos}, \bibinfo{person}{Matthias Zwicker}, \bibinfo{person}{Aljosa
  Smolic}, {and} \bibinfo{person}{Markus Gross}.}
  \bibinfo{year}{2014}\natexlab{}.
\newblock \showarticletitle{Optimizing stereo-to-multiview conversion for
  autostereoscopic displays}. In \bibinfo{booktitle}{\emph{Computer graphics
  forum}}, Vol.~\bibinfo{volume}{33}. Wiley Online Library,
  \bibinfo{pages}{63--72}.
\newblock


\bibitem[Chaurasia et~al\mbox{.}(2020)]%
        {chaurasia2020passthrough+}
\bibfield{author}{\bibinfo{person}{Gaurav Chaurasia}, \bibinfo{person}{Arthur
  Nieuwoudt}, \bibinfo{person}{Alexandru-Eugen Ichim}, \bibinfo{person}{Richard
  Szeliski}, {and} \bibinfo{person}{Alexander Sorkine-Hornung}.}
  \bibinfo{year}{2020}\natexlab{}.
\newblock \showarticletitle{Passthrough+ Real-time Stereoscopic View Synthesis
  for Mobile Mixed Reality}.
\newblock \bibinfo{journal}{\emph{Proceedings of the ACM on Computer Graphics
  and Interactive Techniques}} \bibinfo{volume}{3}, \bibinfo{number}{1}
  (\bibinfo{year}{2020}), \bibinfo{pages}{1--17}.
\newblock


\bibitem[Chen and Williams(1993)]%
        {chen1993view}
\bibfield{author}{\bibinfo{person}{Shenchang~Eric Chen} {and}
  \bibinfo{person}{Lance Williams}.} \bibinfo{year}{1993}\natexlab{}.
\newblock \showarticletitle{View interpolation for image synthesis}. In
  \bibinfo{booktitle}{\emph{Proceedings of the 20th annual conference on
  Computer graphics and interactive techniques}}. \bibinfo{pages}{279--288}.
\newblock


\bibitem[Didyk et~al\mbox{.}(2013)]%
        {didyk2013joint}
\bibfield{author}{\bibinfo{person}{Piotr Didyk}, \bibinfo{person}{Pitchaya
  Sitthi-Amorn}, \bibinfo{person}{William Freeman}, \bibinfo{person}{Fr{\'e}do
  Durand}, {and} \bibinfo{person}{Wojciech Matusik}.}
  \bibinfo{year}{2013}\natexlab{}.
\newblock \showarticletitle{Joint view expansion and filtering for
  automultiscopic 3D displays}.
\newblock \bibinfo{journal}{\emph{ACM Transactions on Graphics (TOG)}}
  \bibinfo{volume}{32}, \bibinfo{number}{6} (\bibinfo{year}{2013}),
  \bibinfo{pages}{1--8}.
\newblock


\bibitem[Flynn et~al\mbox{.}(2019)]%
        {flynn2019deepview}
\bibfield{author}{\bibinfo{person}{John Flynn}, \bibinfo{person}{Michael
  Broxton}, \bibinfo{person}{Paul Debevec}, \bibinfo{person}{Matthew DuVall},
  \bibinfo{person}{Graham Fyffe}, \bibinfo{person}{Ryan Overbeck},
  \bibinfo{person}{Noah Snavely}, {and} \bibinfo{person}{Richard Tucker}.}
  \bibinfo{year}{2019}\natexlab{}.
\newblock \showarticletitle{Deepview: View synthesis with learned gradient
  descent}. In \bibinfo{booktitle}{\emph{Proceedings of the IEEE/CVF Conference
  on Computer Vision and Pattern Recognition}}. \bibinfo{pages}{2367--2376}.
\newblock


\bibitem[Gao et~al\mbox{.}(2021)]%
        {gao2021dynamic}
\bibfield{author}{\bibinfo{person}{Chen Gao}, \bibinfo{person}{Ayush Saraf},
  \bibinfo{person}{Johannes Kopf}, {and} \bibinfo{person}{Jia-Bin Huang}.}
  \bibinfo{year}{2021}\natexlab{}.
\newblock \showarticletitle{Dynamic View Synthesis from Dynamic Monocular
  Video}.
\newblock \bibinfo{journal}{\emph{arXiv preprint arXiv:2105.06468}}
  (\bibinfo{year}{2021}).
\newblock


\bibitem[Garbin et~al\mbox{.}(2021)]%
        {garbin2021fastnerf}
\bibfield{author}{\bibinfo{person}{Stephan~J Garbin}, \bibinfo{person}{Marek
  Kowalski}, \bibinfo{person}{Matthew Johnson}, \bibinfo{person}{Jamie
  Shotton}, {and} \bibinfo{person}{Julien Valentin}.}
  \bibinfo{year}{2021}\natexlab{}.
\newblock \showarticletitle{Fastnerf: High-fidelity neural rendering at
  200fps}. In \bibinfo{booktitle}{\emph{Proceedings of the IEEE/CVF
  International Conference on Computer Vision}}. \bibinfo{pages}{14346--14355}.
\newblock


\bibitem[Gortler et~al\mbox{.}(1996)]%
        {gortler1996lumigraph}
\bibfield{author}{\bibinfo{person}{Steven~J Gortler}, \bibinfo{person}{Radek
  Grzeszczuk}, \bibinfo{person}{Richard Szeliski}, {and}
  \bibinfo{person}{Michael~F Cohen}.} \bibinfo{year}{1996}\natexlab{}.
\newblock \showarticletitle{The lumigraph}. In
  \bibinfo{booktitle}{\emph{Proceedings of the 23rd annual conference on
  Computer graphics and interactive techniques}}. \bibinfo{pages}{43--54}.
\newblock


\bibitem[Guo et~al\mbox{.}(2022)]%
        {guo2022fast}
\bibfield{author}{\bibinfo{person}{Pengsheng Guo},
  \bibinfo{person}{Miguel~Angel Bautista}, \bibinfo{person}{Alex Colburn},
  \bibinfo{person}{Liang Yang}, \bibinfo{person}{Daniel Ulbricht},
  \bibinfo{person}{Joshua~M Susskind}, {and} \bibinfo{person}{Qi Shan}.}
  \bibinfo{year}{2022}\natexlab{}.
\newblock \showarticletitle{Fast and Explicit Neural View Synthesis}. In
  \bibinfo{booktitle}{\emph{Proceedings of the IEEE/CVF Winter Conference on
  Applications of Computer Vision}}. \bibinfo{pages}{3791--3800}.
\newblock


\bibitem[Hedman et~al\mbox{.}(2017)]%
        {hedman2017casual}
\bibfield{author}{\bibinfo{person}{Peter Hedman}, \bibinfo{person}{Suhib
  Alsisan}, \bibinfo{person}{Richard Szeliski}, {and} \bibinfo{person}{Johannes
  Kopf}.} \bibinfo{year}{2017}\natexlab{}.
\newblock \showarticletitle{Casual 3D photography}.
\newblock \bibinfo{journal}{\emph{ACM Transactions on Graphics (TOG)}}
  \bibinfo{volume}{36}, \bibinfo{number}{6} (\bibinfo{year}{2017}),
  \bibinfo{pages}{1--15}.
\newblock


\bibitem[Hedman et~al\mbox{.}(2021)]%
        {hedman2021baking}
\bibfield{author}{\bibinfo{person}{Peter Hedman}, \bibinfo{person}{Pratul~P
  Srinivasan}, \bibinfo{person}{Ben Mildenhall}, \bibinfo{person}{Jonathan~T
  Barron}, {and} \bibinfo{person}{Paul Debevec}.}
  \bibinfo{year}{2021}\natexlab{}.
\newblock \showarticletitle{Baking neural radiance fields for real-time view
  synthesis}. In \bibinfo{booktitle}{\emph{Proceedings of the IEEE/CVF
  International Conference on Computer Vision}}. \bibinfo{pages}{5875--5884}.
\newblock


\bibitem[Jain et~al\mbox{.}(2021)]%
        {jain2021putting}
\bibfield{author}{\bibinfo{person}{Ajay Jain}, \bibinfo{person}{Matthew
  Tancik}, {and} \bibinfo{person}{Pieter Abbeel}.}
  \bibinfo{year}{2021}\natexlab{}.
\newblock \showarticletitle{Putting nerf on a diet: Semantically consistent
  few-shot view synthesis}. In \bibinfo{booktitle}{\emph{Proceedings of the
  IEEE/CVF International Conference on Computer Vision}}.
  \bibinfo{pages}{5885--5894}.
\newblock


\bibitem[Kalantari et~al\mbox{.}(2016)]%
        {kalantari2016learning}
\bibfield{author}{\bibinfo{person}{Nima~Khademi Kalantari},
  \bibinfo{person}{Ting-Chun Wang}, {and} \bibinfo{person}{Ravi Ramamoorthi}.}
  \bibinfo{year}{2016}\natexlab{}.
\newblock \showarticletitle{Learning-based view synthesis for light field
  cameras}.
\newblock \bibinfo{journal}{\emph{ACM Transactions on Graphics (TOG)}}
  \bibinfo{volume}{35}, \bibinfo{number}{6} (\bibinfo{year}{2016}),
  \bibinfo{pages}{1--10}.
\newblock


\bibitem[Kopf et~al\mbox{.}(2020)]%
        {kopf2020one}
\bibfield{author}{\bibinfo{person}{Johannes Kopf}, \bibinfo{person}{Kevin
  Matzen}, \bibinfo{person}{Suhib Alsisan}, \bibinfo{person}{Ocean Quigley},
  \bibinfo{person}{Francis Ge}, \bibinfo{person}{Yangming Chong},
  \bibinfo{person}{Josh Patterson}, \bibinfo{person}{Jan-Michael Frahm},
  \bibinfo{person}{Shu Wu}, \bibinfo{person}{Matthew Yu}, {et~al\mbox{.}}}
  \bibinfo{year}{2020}\natexlab{}.
\newblock \showarticletitle{One shot 3d photography}.
\newblock \bibinfo{journal}{\emph{ACM Transactions on Graphics (TOG)}}
  \bibinfo{volume}{39}, \bibinfo{number}{4} (\bibinfo{year}{2020}),
  \bibinfo{pages}{76--1}.
\newblock


\bibitem[Krajancich et~al\mbox{.}(2020)]%
        {Krajancich2020gcr}
\bibfield{author}{\bibinfo{person}{Brooke Krajancich}, \bibinfo{person}{Petr
  Kellnhofer}, {and} \bibinfo{person}{Gordon Wetzstein}.}
  \bibinfo{year}{2020}\natexlab{}.
\newblock \showarticletitle{Optimizing Depth Perception in Virtual and
  Augmented Reality through Gaze-Contingent Stereo Rendering}.
\newblock \bibinfo{journal}{\emph{ACM Trans. Graph.}} \bibinfo{volume}{39},
  \bibinfo{number}{6}, Article \bibinfo{articleno}{269} (\bibinfo{date}{nov}
  \bibinfo{year}{2020}), \bibinfo{numpages}{10}~pages.
\newblock


\bibitem[Levoy and Hanrahan(1996)]%
        {levoy1996light}
\bibfield{author}{\bibinfo{person}{Marc Levoy} {and} \bibinfo{person}{Pat
  Hanrahan}.} \bibinfo{year}{1996}\natexlab{}.
\newblock \showarticletitle{Light field rendering}. In
  \bibinfo{booktitle}{\emph{Proceedings of the 23rd annual conference on
  Computer graphics and interactive techniques}}. \bibinfo{pages}{31--42}.
\newblock


\bibitem[Lipson et~al\mbox{.}(2021)]%
        {lipson2021raft}
\bibfield{author}{\bibinfo{person}{Lahav Lipson}, \bibinfo{person}{Zachary
  Teed}, {and} \bibinfo{person}{Jia Deng}.} \bibinfo{year}{2021}\natexlab{}.
\newblock \showarticletitle{Raft-stereo: Multilevel recurrent field transforms
  for stereo matching}.
\newblock \bibinfo{journal}{\emph{arXiv preprint arXiv:2109.07547}}
  (\bibinfo{year}{2021}).
\newblock


\bibitem[Liu et~al\mbox{.}(2018)]%
        {liu2018image}
\bibfield{author}{\bibinfo{person}{Guilin Liu}, \bibinfo{person}{Fitsum~A
  Reda}, \bibinfo{person}{Kevin~J Shih}, \bibinfo{person}{Ting-Chun Wang},
  \bibinfo{person}{Andrew Tao}, {and} \bibinfo{person}{Bryan Catanzaro}.}
  \bibinfo{year}{2018}\natexlab{}.
\newblock \showarticletitle{Image inpainting for irregular holes using partial
  convolutions}. In \bibinfo{booktitle}{\emph{Proceedings of the European
  Conference on Computer Vision (ECCV)}}. \bibinfo{pages}{85--100}.
\newblock


\bibitem[Liu et~al\mbox{.}(2020)]%
        {liu2020neural}
\bibfield{author}{\bibinfo{person}{Lingjie Liu}, \bibinfo{person}{Jiatao Gu},
  \bibinfo{person}{Kyaw~Zaw Lin}, \bibinfo{person}{Tat-Seng Chua}, {and}
  \bibinfo{person}{Christian Theobalt}.} \bibinfo{year}{2020}\natexlab{}.
\newblock \showarticletitle{Neural sparse voxel fields}.
\newblock \bibinfo{journal}{\emph{arXiv preprint arXiv:2007.11571}}
  (\bibinfo{year}{2020}).
\newblock


\bibitem[Maimone and Wang(2020)]%
        {maimone2020holocake}
\bibfield{author}{\bibinfo{person}{Andrew Maimone} {and}
  \bibinfo{person}{Junren Wang}.} \bibinfo{year}{2020}\natexlab{}.
\newblock \showarticletitle{Holographic Optics for Thin and Lightweight Virtual
  Reality}.
\newblock \bibinfo{journal}{\emph{ACM Trans. Graph.}} \bibinfo{volume}{39},
  \bibinfo{number}{4}, Article \bibinfo{articleno}{67} (\bibinfo{date}{jul}
  \bibinfo{year}{2020}), \bibinfo{numpages}{14}~pages.
\newblock


\bibitem[Martin-Brualla et~al\mbox{.}(2018)]%
        {martin2018lookingood}
\bibfield{author}{\bibinfo{person}{Ricardo Martin-Brualla},
  \bibinfo{person}{Rohit Pandey}, \bibinfo{person}{Shuoran Yang},
  \bibinfo{person}{Pavel Pidlypenskyi}, \bibinfo{person}{Jonathan Taylor},
  \bibinfo{person}{Julien Valentin}, \bibinfo{person}{Sameh Khamis},
  \bibinfo{person}{Philip Davidson}, \bibinfo{person}{Anastasia Tkach},
  \bibinfo{person}{Peter Lincoln}, {et~al\mbox{.}}}
  \bibinfo{year}{2018}\natexlab{}.
\newblock \showarticletitle{LookinGood: enhancing performance capture with
  real-time neural re-rendering}.
\newblock \bibinfo{journal}{\emph{ACM Transactions on Graphics (TOG)}}
  \bibinfo{volume}{37}, \bibinfo{number}{6} (\bibinfo{year}{2018}),
  \bibinfo{pages}{1--14}.
\newblock


\bibitem[Mildenhall et~al\mbox{.}(2019)]%
        {mildenhall2019local}
\bibfield{author}{\bibinfo{person}{Ben Mildenhall}, \bibinfo{person}{Pratul~P
  Srinivasan}, \bibinfo{person}{Rodrigo Ortiz-Cayon},
  \bibinfo{person}{Nima~Khademi Kalantari}, \bibinfo{person}{Ravi Ramamoorthi},
  \bibinfo{person}{Ren Ng}, {and} \bibinfo{person}{Abhishek Kar}.}
  \bibinfo{year}{2019}\natexlab{}.
\newblock \showarticletitle{Local light field fusion: Practical view synthesis
  with prescriptive sampling guidelines}.
\newblock \bibinfo{journal}{\emph{ACM Transactions on Graphics (TOG)}}
  \bibinfo{volume}{38}, \bibinfo{number}{4} (\bibinfo{year}{2019}),
  \bibinfo{pages}{1--14}.
\newblock


\bibitem[Mildenhall et~al\mbox{.}(2020)]%
        {mildenhall2020nerf}
\bibfield{author}{\bibinfo{person}{Ben Mildenhall}, \bibinfo{person}{Pratul~P
  Srinivasan}, \bibinfo{person}{Matthew Tancik}, \bibinfo{person}{Jonathan~T
  Barron}, \bibinfo{person}{Ravi Ramamoorthi}, {and} \bibinfo{person}{Ren Ng}.}
  \bibinfo{year}{2020}\natexlab{}.
\newblock \showarticletitle{Nerf: Representing scenes as neural radiance fields
  for view synthesis}. In \bibinfo{booktitle}{\emph{European conference on
  computer vision}}. Springer, \bibinfo{pages}{405--421}.
\newblock


\bibitem[Niklaus and Liu(2020)]%
        {niklaus2020softmax}
\bibfield{author}{\bibinfo{person}{Simon Niklaus} {and} \bibinfo{person}{Feng
  Liu}.} \bibinfo{year}{2020}\natexlab{}.
\newblock \showarticletitle{Softmax splatting for video frame interpolation}.
  In \bibinfo{booktitle}{\emph{Proceedings of the IEEE/CVF Conference on
  Computer Vision and Pattern Recognition}}. \bibinfo{pages}{5437--5446}.
\newblock


\bibitem[Oculus(2016)]%
        {asw2}
\bibfield{author}{\bibinfo{person}{Oculus}.} \bibinfo{year}{2016}\natexlab{}.
\newblock \bibinfo{booktitle}{\emph{asynchronous spacewarp}}.
\newblock
\urldef\tempurl%
\url{https://www.oculus.com/blog/introducing-asw-2-point-0-better-accuracy-lower-latency/}
\showURL{%
\tempurl}


\bibitem[Pumarola et~al\mbox{.}(2021)]%
        {pumarola2021d}
\bibfield{author}{\bibinfo{person}{Albert Pumarola}, \bibinfo{person}{Enric
  Corona}, \bibinfo{person}{Gerard Pons-Moll}, {and} \bibinfo{person}{Francesc
  Moreno-Noguer}.} \bibinfo{year}{2021}\natexlab{}.
\newblock \showarticletitle{D-nerf: Neural radiance fields for dynamic scenes}.
  In \bibinfo{booktitle}{\emph{Proceedings of the IEEE/CVF Conference on
  Computer Vision and Pattern Recognition}}. \bibinfo{pages}{10318--10327}.
\newblock


\bibitem[Reiser et~al\mbox{.}(2021)]%
        {reiser2021kilonerf}
\bibfield{author}{\bibinfo{person}{Christian Reiser}, \bibinfo{person}{Songyou
  Peng}, \bibinfo{person}{Yiyi Liao}, {and} \bibinfo{person}{Andreas Geiger}.}
  \bibinfo{year}{2021}\natexlab{}.
\newblock \showarticletitle{KiloNeRF: Speeding up Neural Radiance Fields with
  Thousands of Tiny MLPs}.
\newblock \bibinfo{journal}{\emph{arXiv preprint arXiv:2103.13744}}
  (\bibinfo{year}{2021}).
\newblock


\bibitem[Shade et~al\mbox{.}(1998)]%
        {shade1998layered}
\bibfield{author}{\bibinfo{person}{Jonathan Shade}, \bibinfo{person}{Steven
  Gortler}, \bibinfo{person}{Li-wei He}, {and} \bibinfo{person}{Richard
  Szeliski}.} \bibinfo{year}{1998}\natexlab{}.
\newblock \showarticletitle{Layered depth images}. In
  \bibinfo{booktitle}{\emph{Proceedings of the 25th annual conference on
  Computer graphics and interactive techniques}}. \bibinfo{pages}{231--242}.
\newblock


\bibitem[Shih et~al\mbox{.}(2020)]%
        {shih3d}
\bibfield{author}{\bibinfo{person}{Meng-Li Shih}, \bibinfo{person}{Shih-Yang
  Su}, \bibinfo{person}{Johannes Kopf}, {and} \bibinfo{person}{Jia-Bin Huang}.}
  \bibinfo{year}{2020}\natexlab{}.
\newblock \showarticletitle{3d photography using context-aware layered depth
  inpainting}. In \bibinfo{booktitle}{\emph{Proceedings of the IEEE Conference
  on Computer Vision and Pattern Recognition}}. \bibinfo{pages}{8028--8038}.
\newblock


\bibitem[Soundararajan and Bovik(2012)]%
        {soundararajan2012video}
\bibfield{author}{\bibinfo{person}{Rajiv Soundararajan} {and}
  \bibinfo{person}{Alan~C Bovik}.} \bibinfo{year}{2012}\natexlab{}.
\newblock \showarticletitle{Video quality assessment by reduced reference
  spatio-temporal entropic differencing}.
\newblock \bibinfo{journal}{\emph{IEEE Transactions on Circuits and Systems for
  Video Technology}} \bibinfo{volume}{23}, \bibinfo{number}{4}
  (\bibinfo{year}{2012}), \bibinfo{pages}{684--694}.
\newblock


\bibitem[Srinivasan et~al\mbox{.}(2019)]%
        {srinivasan2019pushing}
\bibfield{author}{\bibinfo{person}{Pratul~P Srinivasan},
  \bibinfo{person}{Richard Tucker}, \bibinfo{person}{Jonathan~T Barron},
  \bibinfo{person}{Ravi Ramamoorthi}, \bibinfo{person}{Ren Ng}, {and}
  \bibinfo{person}{Noah Snavely}.} \bibinfo{year}{2019}\natexlab{}.
\newblock \showarticletitle{Pushing the boundaries of view extrapolation with
  multiplane images}. In \bibinfo{booktitle}{\emph{Proceedings of the IEEE/CVF
  Conference on Computer Vision and Pattern Recognition}}.
  \bibinfo{pages}{175--184}.
\newblock


\bibitem[Wang et~al\mbox{.}(2004)]%
        {wang2004image}
\bibfield{author}{\bibinfo{person}{Zhou Wang}, \bibinfo{person}{Alan~C Bovik},
  \bibinfo{person}{Hamid~R Sheikh}, \bibinfo{person}{Eero~P Simoncelli},
  {et~al\mbox{.}}} \bibinfo{year}{2004}\natexlab{}.
\newblock \showarticletitle{Image quality assessment: from error visibility to
  structural similarity}.
\newblock \bibinfo{journal}{\emph{IEEE Transactions on Image Processing}}
  \bibinfo{volume}{13}, \bibinfo{number}{4} (\bibinfo{year}{2004}),
  \bibinfo{pages}{600--612}.
\newblock


\bibitem[Wiles et~al\mbox{.}(2020)]%
        {wiles2020synsin}
\bibfield{author}{\bibinfo{person}{Olivia Wiles}, \bibinfo{person}{Georgia
  Gkioxari}, \bibinfo{person}{Richard Szeliski}, {and} \bibinfo{person}{Justin
  Johnson}.} \bibinfo{year}{2020}\natexlab{}.
\newblock \showarticletitle{Synsin: End-to-end view synthesis from a single
  image}. In \bibinfo{booktitle}{\emph{Proceedings of the IEEE/CVF Conference
  on Computer Vision and Pattern Recognition}}. \bibinfo{pages}{7467--7477}.
\newblock


\bibitem[Xiao et~al\mbox{.}(2018)]%
        {xiao2018deepfocus}
\bibfield{author}{\bibinfo{person}{Lei Xiao}, \bibinfo{person}{Anton
  Kaplanyan}, \bibinfo{person}{Alexander Fix}, \bibinfo{person}{Matthew
  Chapman}, {and} \bibinfo{person}{Douglas Lanman}.}
  \bibinfo{year}{2018}\natexlab{}.
\newblock \showarticletitle{DeepFocus: learned image synthesis for
  computational displays}.
\newblock \bibinfo{journal}{\emph{ACM Transactions on Graphics (TOG)}}
  \bibinfo{volume}{37}, \bibinfo{number}{6} (\bibinfo{year}{2018}),
  \bibinfo{pages}{1--13}.
\newblock


\bibitem[Yu et~al\mbox{.}(2021a)]%
        {yu2021plenoctrees}
\bibfield{author}{\bibinfo{person}{Alex Yu}, \bibinfo{person}{Ruilong Li},
  \bibinfo{person}{Matthew Tancik}, \bibinfo{person}{Hao Li},
  \bibinfo{person}{Ren Ng}, {and} \bibinfo{person}{Angjoo Kanazawa}.}
  \bibinfo{year}{2021}\natexlab{a}.
\newblock \showarticletitle{Plenoctrees for real-time rendering of neural
  radiance fields}.
\newblock \bibinfo{journal}{\emph{arXiv preprint arXiv:2103.14024}}
  (\bibinfo{year}{2021}).
\newblock


\bibitem[Yu et~al\mbox{.}(2021b)]%
        {yu2021pixelnerf}
\bibfield{author}{\bibinfo{person}{Alex Yu}, \bibinfo{person}{Vickie Ye},
  \bibinfo{person}{Matthew Tancik}, {and} \bibinfo{person}{Angjoo Kanazawa}.}
  \bibinfo{year}{2021}\natexlab{b}.
\newblock \showarticletitle{pixelnerf: Neural radiance fields from one or few
  images}. In \bibinfo{booktitle}{\emph{Proceedings of the IEEE/CVF Conference
  on Computer Vision and Pattern Recognition}}. \bibinfo{pages}{4578--4587}.
\newblock


\bibitem[Zhang et~al\mbox{.}(2020)]%
        {zhang2020nerf++}
\bibfield{author}{\bibinfo{person}{Kai Zhang}, \bibinfo{person}{Gernot
  Riegler}, \bibinfo{person}{Noah Snavely}, {and} \bibinfo{person}{Vladlen
  Koltun}.} \bibinfo{year}{2020}\natexlab{}.
\newblock \showarticletitle{Nerf++: Analyzing and improving neural radiance
  fields}.
\newblock \bibinfo{journal}{\emph{arXiv preprint arXiv:2010.07492}}
  (\bibinfo{year}{2020}).
\newblock


\bibitem[Zhou et~al\mbox{.}(2018)]%
        {zhou2018stereo}
\bibfield{author}{\bibinfo{person}{Tinghui Zhou}, \bibinfo{person}{Richard
  Tucker}, \bibinfo{person}{John Flynn}, \bibinfo{person}{Graham Fyffe}, {and}
  \bibinfo{person}{Noah Snavely}.} \bibinfo{year}{2018}\natexlab{}.
\newblock \showarticletitle{Stereo magnification: Learning view synthesis using
  multiplane images}.
\newblock \bibinfo{journal}{\emph{arXiv preprint arXiv:1805.09817}}
  (\bibinfo{year}{2018}).
\newblock


\bibitem[Zitnick et~al\mbox{.}(2004)]%
        {zitnick2004high}
\bibfield{author}{\bibinfo{person}{C~Lawrence Zitnick},
  \bibinfo{person}{Sing~Bing Kang}, \bibinfo{person}{Matthew Uyttendaele},
  \bibinfo{person}{Simon Winder}, {and} \bibinfo{person}{Richard Szeliski}.}
  \bibinfo{year}{2004}\natexlab{}.
\newblock \showarticletitle{High-quality video view interpolation using a
  layered representation}.
\newblock \bibinfo{journal}{\emph{ACM transactions on graphics (TOG)}}
  \bibinfo{volume}{23}, \bibinfo{number}{3} (\bibinfo{year}{2004}),
  \bibinfo{pages}{600--608}.
\newblock


\end{thebibliography}
